\documentclass[authoryear,preprint]{elsarticle}

\usepackage{caption}
\usepackage{natbib}
\usepackage{appendix}
\usepackage{amsmath}
\usepackage{amssymb}
\usepackage{graphicx}
\usepackage{algorithmic}
\usepackage{hyperref}
\usepackage{url}
\usepackage{multirow}
\usepackage{booktabs}
\usepackage{setspace}
\usepackage{rotating}
\usepackage{hyperref}
\usepackage{algorithmic}
\usepackage[vlined,lined,boxed,commentsnumbered]{algorithm2e}

\newcommand\fnurl[2]{%
	\href{#2}{#1}\footnote{\url{#2}}%
}

\journal{Expert Systems with Applications}

\begin{document}

\begin{frontmatter}

\title{Recurring Concept Meta-learning for Evolving Data Streams}

\author{Robert Anderson\corref{cor1}}
\ead{rand079@aucklanduni.ac.nz}
\author{Yun Sing Koh}
\ead{ykoh@cs.auckland.ac.nz}
\author{Gillian Dobbie}
\ead{gill@cs.auckland.ac.nz}
\address{School of Computer Science, University of Auckland, Auckland, New Zealand}

\author{Albert Bifet}
\address{Telecom ParisTech, 46 Rue Barrault, Paris, France}
\ead{abifet@gmail.com}

\cortext[cor1]{Corresponding author}



\begin{abstract}
\label{sec:abstract}
	
When concept drift is detected during classification in a data stream, a common remedy is to retrain a framework's classifier. However, this loses useful information if the classifier has learnt the current concept well, and this concept will recur again in the future. Some frameworks retain and reuse classifiers, but it can be time-consuming to select an appropriate classifier to reuse. These frameworks rarely match the accuracy of state-of-the-art ensemble approaches. For many data stream tasks, speed is important: fast, accurate frameworks are needed for time-dependent applications. We propose the Enhanced Concept Profiling Framework (ECPF), which aims to recognise recurring concepts and reuse a classifier trained previously, enabling accurate classification immediately following a drift. The novelty of ECPF is in how it uses similarity of classifications on new data, between a new classifier and existing classifiers, to quickly identify the best classifier to reuse. It always trains both a new classifier and a reused classifier, and retains the more accurate classifier when concept drift occurs. Finally, it creates a copy of reused classifiers, so a classifier well-suited for a recurring concept will not be impacted by being trained on a different concept. In our experiments, ECPF classifies significantly more accurately than a state-of-the-art classifier reuse framework (Diversity Pool) and a state-of-the-art ensemble technique (Adaptive Random Forest) on synthetic datasets with recurring concepts. It classifies real-world datasets five times faster than Diversity Pool, and six times faster than Adaptive Random Forest and is not significantly less accurate than either.

\end{abstract}


\end{frontmatter}

\section{Introduction}
\label{sec:introduction}

Learning involves remembering patterns from our past to understand situations we encounter in our future. The more we remember, the better chance we have of applying prior learning to future situations. When classifying instances in data streams however, learning systems cannot store every instance seen. In fact, as the current concept (\emph{i.e.} the underlying data distribution) in streams can change over time (\emph{concept drift}), learning systems may benefit from only being trained on recent data from these evolving streams. Concept drift handling techniques \citep{tsy2004} apply learning from relevant and current data while avoiding bias from outdated data. A common approach is to retrain a classifier whenever a change is detected in the underlying data. However, if a data stream reverts to a previous concept, the learning system then needs to relearn the concept. If we can recognise recurring concepts, we may perform faster and more accurate classification by reverting to classifiers previously trained on those concepts \citep{wid1996}. 

Repeated patterns are regularly observed in the real world: seasons; boom-bust periods in financial markets; rush-hour in traffic flows; and battery states of sensors, for example. However, techniques that try to take advantage of recurring concepts generally use time-consuming testing to identify recurring concepts. For example, RCD \citep{goncalves2013} compares incoming data using kNN while Diversity Pool \citep{chi2018} regularly measures entropy between classifiers. Ensemble approaches, meanwhile, must maintain a large set of classifiers. These approaches typically have high runtime. Some data stream applications operate in extremely time-sensitive environments, such as flood warning systems or automated trading platforms where classification delay is costly. These applications would benefit from recognising and exploiting prior learning on recurring concepts, but the runtime of existing approaches reduce their value. This work addresses the challenge of creating a framework that can exploit recurring concepts while running faster and as accurately as existing approaches. 

In this paper, we present Enhanced CPF (ECPF) which can match the classification accuracy of state-of-the-art approaches on datasets with recurring drift while running much faster. Its novelty is in how it uses similarity of classifications on new data, between a new classifier and existing classifiers, to quickly identify the best classifier to reuse.It creates copies of reused classifiers rather than reusing the original, allowing the technique to vary an existing classifier without losing the original. The classifier variant may be better suited to a distinct concept than the original, and ECPF will now have both to choose between. ECPF also compares reused classifiers with new classifiers trained since the last drift, and retains the more accurate of the two when concept drift occurs.

Consider a sensor relay that aims to deliver flood warnings based on a classifier that uses rain measurements as input. Fast flood warnings are critical to prevent damage. In \emph{dry} periods, recent rainfall is less likely to cause flooding than in \emph{wet} periods. When deciding upon flooding risk, the relay must use a fast classifier that will consider current weather patterns: the cost of slow processing could be running out of power and failing to deliver any warning. The classification approach also needs to adapt to different concepts. ECPF can retain classifiers from the \emph{wet} period and reuse them when the concept recurs after a \emph{dry} period. It can copy and vary a root \emph{wet} classifier into multiple other specific concepts \emph{e.g.} \emph{periodic storms} versus \emph{constant drizzle} classifiers. It can harness prior learning to deliver flood warnings faster and at least as accurately as other state-of-the-art approaches.

\begin{figure}[hbtp]
	\begin{center}
		\includegraphics[width=\textwidth]{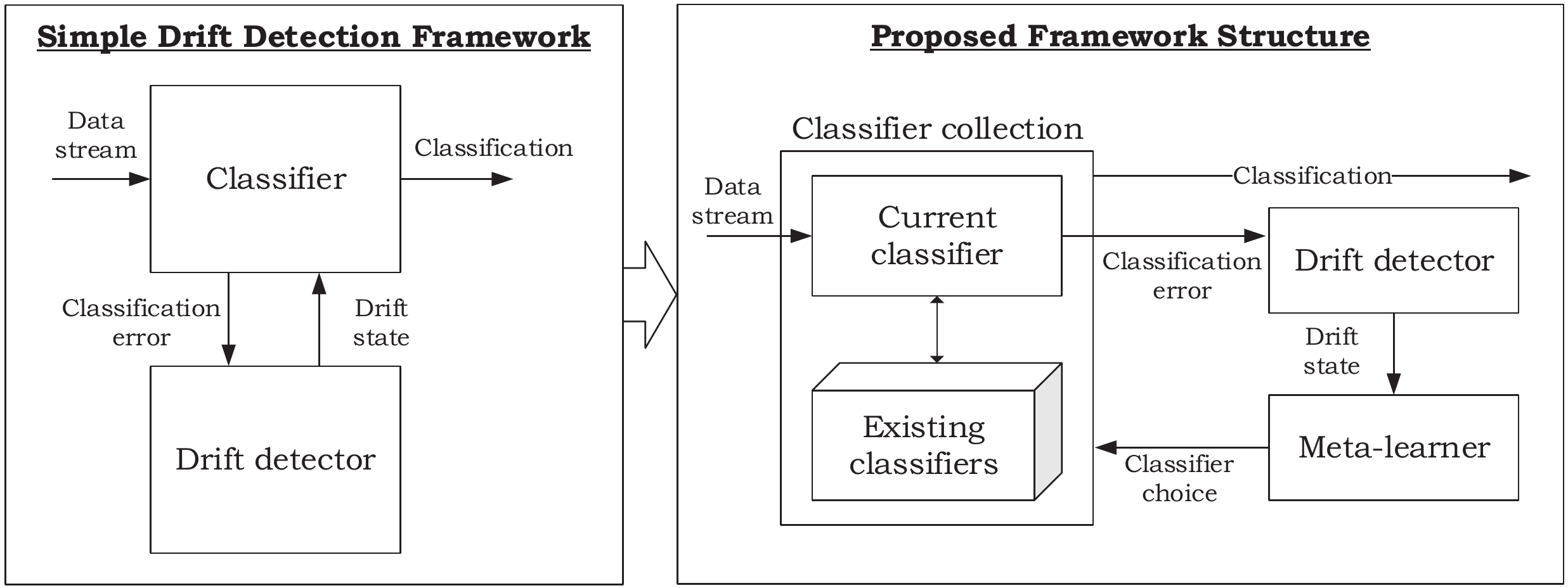}
	\end{center}
	\caption{A simple drift detection framework and our proposed meta-learning framework structure with classifier reuse}
	\label{intro_framework}
\end{figure}

As per Fig. \ref{intro_framework}, our proposed technique classifies instances in a stream over time. When a drift detector signals drift, our meta-learner saves the existing classifier and searches the collection of retained classifiers for one better suited to the new stream concept. This involves selecting an existing classifier with similar classifying behaviour to a new classifier on instances that have arrived after the drift. ECPF creates a new classifier and selects a classifier to reuse, and uses both for classification, delivering the classification of the more accurate classifier of the two. We reduce memory overhead by regularly comparing behaviour of our classifiers: if there are two classifiers that classify similarly as measured by conceptual equivalence \citep{yan2006}, we use one classifier to \emph{represent} both. We also implement a \emph{fading} mechanism to constrain the number of classifiers; a points-based system is used to retain classifiers that are recent or frequently used. Through observing reuse patterns, we can understand how patterns recur in our stream. Our contribution is a meta-learning framework that can utilise observed classifier behaviour over time to recognise when a concept has recurred and leverage classifiers trained previously, using only information from the data stream. ECPF achieves classification accuracy that is comparable to a state-of-the-art ensemble technique and better than a state-of-the-art classifier reuse technique while running at least five times faster than either. ECPF achieves substantially more accurate classifications on real-world benchmark datasets than CPF \citep{and2016}. 

In the next section, we discuss work related to ECPF. In Section 3, we detail our proposed framework, ECPF. In Section 4, we show results for experiments to highlight ECPF's performance compared to recent and state-of-the-art stream classification frameworks. First, we show the contribution it provides in terms of speed and accuracy against current state-of-the-art techniques; we then prove its robustness in the presence of datasets varied in terms of drift period, class balance, noise and with different classifiers; we finally examine the behaviour of our techniques in the presence of different drift detectors and varied parameters. We conclude by summarising our findings and discussing future developments for ECPF.

\section{Related work}
\label{sec:related_work}

In this section, we discuss previous work that has informed this approach. For clarity, we have divided this previous work into four subsections: classification in data streams; detecting concept drift; handling recurring concepts; and the original Concept Profiling Framework.

\subsection{Classification in data streams}

When we discuss data streams, what are we referring to? \citet{gam2010} discusses traditional approaches of data analysis, with finite training sets, static classifiers and stationary distributions. These problems can be easily scoped since researchers can understand the data to be considered when designing the analysis. In contrast, data stream analysis deals with high-speed and potentially infinite data received over time. The underlying distribution may change over the life of the stream. To be able to handle these characteristics, data should be transient: analysed and discarded. Models should be incremental to complement this - they should be able to be updated and should be able to learn from data as it arrives. \citet{gab2005} discuss the challenges in storage and processing of data streams. To handle a potentially unlimited amount of data, we need algorithms that can scale to large amounts of data without continuously demanding extra resources and that can produce results in a timely fashion suitable to the problem domain. Where an underlying stream changes in nature, techniques need to account for this and adjust their analysis appropriately.

\citet{gam2014} provide an overview of how data streams evolve and solutions for handling classification when they do. They define two distinct types of concept drift: real concept drift occurs when the explanatory (predictor) variables’ relationship with the response (class) variable changes; and virtual drift occurs when the balance of response classes in the incoming data changes. Either type of drift can occur in a drifting stream so for a classifier to be effective, it must be able to handle both possibilities.

Decision trees have proven to work well in data stream classification. Very Fast Decision Trees (VFDTs), as introduced by \citet{dom2000} are especially suitable for data stream classification problems. This is an approach to decision tree building that considers each instance as it is passed to the classifier. It achieves a constant time and memory relationship with the number of instances seen, and is guaranteed to achieve performance similar to a conventional classifier. It does this by limiting the number of examples required to be seen at any node using the Hoeffding bound \citep{hoe1963}, which gives a lower limit for the number of instances that need to be considered before being representative of the overall dataset with probability of one minus a given $\delta$ parameter. The Hoeffding bound is independent of the probability distribution generating instances, so does not rely on the nature of the source data to be applicable. However, this type of classifier cannot easily adapt when a stream has underlying real concept drift. It is difficult for a single tree to know when to discard old examples and base its classifications purely off new instances seen. The CVFDT (Concept-adapting VFDT) proposed by  \citet{hul2001} has a mechanism that grows new sub-trees when old splits no longer pass the test applied by the Hoeffding bound. If the new sub-tree is more accurate than the old sub-tree, then it will replace the old sub-tree, and the tree will better classify new data seen. This approach cannot always account for recurrent drift, as classifiers adapt to recent concepts and may no longer classify well on past concepts. 

\subsection{Handling concept drift with drift-detectors}


Instead of a single adapting classifier, some approaches such as our proposed framework, use a drift-detection mechanism. These try to detect underlying concept drift, so a learning framework can take a corrective action. DDM (Drift Detection Mechanism) \citep{gam2004} monitors the error rate of a classifier. When the mean error rate rises above thresholds based on the minimum error rate seen, it signals a warning or drift. Over long periods, DDM can become insensitive as it is based on the average error rate since the last drift. RDDM (Reactive Drift Detective Measure) \cite{bar2017} addresses this issue by calculating error rate over a smaller number of instances. \citet{fri2015} propose two alternate approaches to drift detection. HDDM-A monitors the average error rate over time, but unlike DDM, it does not assume a parametric distribution for error rate. By reformulating the Hoeffding inequality, it can track the moving average error rate and accurately estimate probability of underlying drift regardless of the true probability distribution of the error rate. HDDM-W extends this approach by reformulating McDiarmid's inequality to use a weighted moving average instead, which is proposed as a good approach to detecting gradual drift. RDDM and HDDM-A were found to be the best drift detectors for improving classification accuracy in a recent, comprehensive survey of drift-detection techniques \cite{bar2018}.


ADWIN \citep{bif2007} is a drift-detector that maintains a variably-sized window of recent input to the drift detector, normally based on classifier error. Windows will be sized as large as possible while recording no statistically significant change in error rate within the most recent window. When the algorithm detects a difference of a given threshold in the average of the error rate between the last window and the one prior, it will signal a drift. This threshold is based upon the Hoeffding Bound and assumes the worst-case in terms of standard deviation of the true error rate, so provides a guaranteed level of certainty for the true chance of drift. This was extended by \citet{hua2014} in their algorithm, SEED, to consider frequency of drifts over time (i.e. \emph{volatility}). This additional information is utilised in their approach to weight a decision on the proof required to signal a drift. MagSeed \citep{che2015} was a further extension to this approach. It seeks to reduce incorrectly detected drifts by adding two warning conditions to the approach that need to be met before a drift is detected: the first is a looser version of the Hoeffding bound used in ADWIN and the other is a change in underlying mean error rate between sliding windows in the stream. Because it is not a change in cumulative error rate, this approach can adapt faster to sudden changes than the one used by DDM. A change cannot be detected until these warning flags are triggered. This algorithm provides a time-efficient approach to drift-detection with guarantees of performance. Most interestingly, it provides a window-based approach with a warning level to signal incoming drift.

\subsection{Handling recurring concepts}

Meta-learning approaches to stream classification such as our proposed work try to learn about the nature of the underlying stream with tools within the framework. For example, the approach proposed by \citet{gam2009} handles recurring concepts by building a ‘referee' classifier alongside every instance a classifier sees, and keeps a collection of these pairs. The referee classifier judges whether its classifier is likely to correctly classify an instance  by learning whether it was correct for previous similar instances. When existing classifiers are not applicable, a new classifier is created. These new parts of a framework can come at a cost to the efficiency of stream analysis. For instance, in this technique, no suggestions are made for constraining the total number of classifiers built over time, risking ever-expanding memory requirements.

\citet{gomes2012} propose using an additional user-specified context stream alongside a data stream. Their approach relates the current context to the current classifier when it is performing well. Particular contexts become associated to classifiers that are useful in those contexts. After drift, a classifier is reused if the context learner feels it fits the current context. Otherwise, a new classifier is created. They use \emph{conceptual equivalence} \citep{yan2006}, which relates similar classifiers through their classifying behaviour. This approach requires an additional context stream which is difficult to select for a problem that is not well understood. This technique was extended by \citet{ang2016}, where a framework was proposed that uses Hidden Markov Models to predict upcoming drift, with previously trained classifiers used to test similarity of past concepts with a present concept. This technique does not run efficiently enough for use in data streams.

\citet{goncalves2013} propose RCD, a framework that maintains a classifier collection. For each classifier, a set of recent training instances are stored to represent its concept. When a drift is detected, a statistical test (kNN) compares the instances in the warning buffer to the training instances for classifiers. If the instances are found to be significantly similar, it uses the corresponding classifier; otherwise, a new one is created. The statistical testing and buffer stores add significant runtime and memory requirements to their approach. Diversity Pool (DP) \citep{chi2018} is a state-of-the-art approach that also maintains a pool of classifiers, reusing select classifiers through the stream. It uses entropy measurements to maximise the difference between classifiers, to improve the chances of having a suitable classifier on drift detection. We have elected to use RCD as a benchmark technique for our approach, as like our work, it takes advantage of collections of classifiers trained prior. RCD measures changes in underlying data rather than in classifier behaviour, making it robust across different types of dataset. We wish for our techniques to be similarly robust.

Ensemble approaches to recurrent learning use multiple classifiers examining a stream, allowing greater chance to find one that functions well at a given time. As a result, they are known to generally outperform single-classifier techniques in terms of accuracy, but to require more time and memory to run. AUE \citep{brz2014} is one example of a current ensemble technique for classifying data streams. It maintains a collection of ten Hoeffding Trees (though other component classifiers can be used), with the weakest performing member regularly being replaced by a new tree, based on performance on recent data. Adaptive Random Forests (ARF) \cite{gom2017} has been shown to be a leading stream classification technique in terms of classification accuracy. This approach builds a diverse set of classifiers, each which can be replaced by a new tree trained on data seen since drift was last detected, according to ADWIN which is run as part of the ensemble.


\subsection{Concept Profiling Framework (CPF)}

The Concept Profiling Framework (CPF) as proposed in \cite{and2016}, recognises recurring concepts through finding past classifiers that classify recent data in a similar fashion to a new classifier. It then reuses these classifiers, allowing accurate classifications without having to relearn the underlying concept. This framework uses a collection of one or more incremental classifiers. One is designated as the current classifier. A drift detector signals warning and drift states. On a warning state, the meta-learner will stop training the current classifier and store instances from the data stream in a buffer of a pre-set size. If a drift state follows, the meta-learner looks for an existing classifier in the collection that classifies the warning buffer accurately to use as the current classifier. If it cannot find one, it will create a new classifier trained on even buffer instances. Every existing classifier in the collection is tested on the buffer, and the one that behaves similarly enough to this new classifier (when tested on odd buffer instances) will be reused; otherwise the new classifier will be trained on odd buffer instances and used instead. Similarity of classifiers' classifying behaviour is stored and compared. Where it is found that classifiers classify similarly to one another, the older classifier will \emph{represent} the newer one.

Through regular reuse and representation of particular classifiers, CPF aims for its classifiers to classify particular concepts very well over time. Frequency of reuse and representation can show patterns of recurrence in the underlying data to understand the underlying concept patterns in the stream. CPF can pair with any classifier that is incremental and performs suitably for a streaming environment. However, it relies on a buffer of a set size of new instances when determining classifiers to reuse. This makes it slower to react to drift than other approaches. A classifier that works well on the concept may no longer work well after being reused on a different concept as the classifier will be changed by new training data. Finally, CPF retains classifiers that have been trained on more instances over those trained on fewer. These may perform well on recurrent concepts similar to those seen prior, but when incorrectly reused, they may not generalise well. CPF is shown to be very effective on synthetic data with clear recurrent drifts, but fails to notably outperform RCD on real-world benchmarks.

\section{Enhanced Concept Profiling Framework}
\label{sec:proposed_frameworks}

In this section, we detail the Enhanced Concept Profiling Framework (ECPF), and how it enhances the design of CPF. By providing more flexible ways to store and reuse previously trained classifiers, ECPF reliably classifies more accurately than CPF on many datasets, and never performs worse. We first describe how ECPF achieves this, through explaining how it classifies new data and manages its classifier collection. We detail \emph{classifier reuse} and \emph{classifier representation}, and when and how they occur in our proposed technique. We define our chosen classifier similarity measure, \emph{conceptual equivalence} and describe the \emph{fading} mechanism used by ECPF. We provide pseudocode for ECPF and describe ECPF's behaviour using an illustrative example.

\begin{figure}
	\begin{center}
		\includegraphics[width=\textwidth]{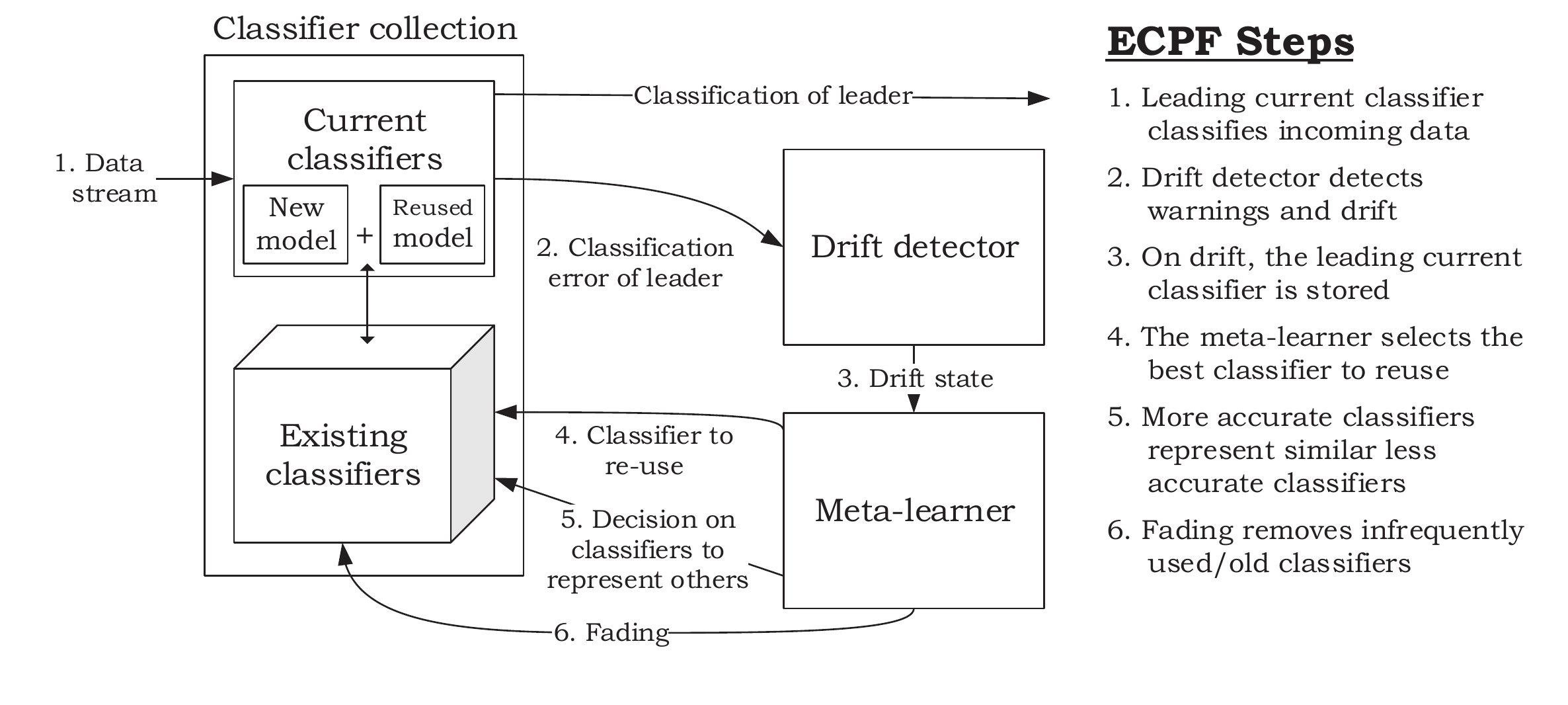}
	\end{center}
	\caption{The Enhanced Concept Profiling Framework}
	\label{ECPF}
\end{figure}

As per Fig \ref{ECPF} and described in Algorithm \ref{ECPFalg}, ECPF constantly maintains a collection of classifiers. Before the first time drift is detected, only a new classifier exists, but afterwards it simultaneously trains two classifiers. The first is a new classifier, trained on the warning buffer \emph{i.e} instances seen between the drift detector's warning threshold and actual drift detection for the prior drift. The second is an existing classifier from the classifier collection that achieves the best accuracy on the warning buffer. Only the classification from the current leading classifier (in terms of accuracy) is returned to the end user (lines 2-3); the classification error from that classifier is considered by the drift detector for the purposes of drift detection (line 4).

When a drift detector signals a warning, classifiers are no longer trained on incoming instances. The instances instead will be inserted into a buffer (line 6). If no drift is detected, ECPF will return to training the two classifiers on new instances (lines 16 and 17). If drift is detected (lines 8 -- 14), the more accurate classifier will be saved to the classifier collection and a new classifier will be trained on the buffer instances. Existing classifiers will be tested on the buffer, and the most accurate will be selected to be reused. This involves copying the classifier, with the copy trained and used in tandem with a new classifier on instances after the drift point. We copy the existing classifier as sometimes it will not be suitable on the new concept. We do not wish to change the original copy of the reused classifier by training it on instances from a new concept, as that may stop it being effective on its original concept. If the reused classifier is appropriate for the new concept, it should classify more accurately than a new classifier, and so we will keep this copy. If the reused classifier is inappropriate for the concept, it will likely classify the new instances poorly and we will retain the new classifier instead. At any given drift point, ECPF will operate at least as accurately as a simple drift detection framework. This is because ECPF always has a new classifier as well a reused classifier. Reusing inappropriate classifiers will not increase overall framework error, as the new classifier will be more accurate and will be used for classifying new instances. We describe classifier \emph{representation} and \emph{fading} (lines 12-13) later in this section.

\LinesNumbered
\begin{algorithm}[hb!]
	\small
	\KwData{Instances $x$ in stream $X$}
	\KwIn{$C$ -- classifier collection; $c_r$ -- reused classifier; $c_n$ -- new classifier; $dd$ -- drift detector; $b$ -- warning buffer; $m$ -- similarity parameter; $f$ -- fade points parameter}
	\small
	\ForEach(){$x$ in $X$}
	{
		$c \leftarrow$ $c_r$ or $c_n$ with greater accuracy since last drift \\
		$h \leftarrow c$.classify($x$) \\
		$dd.$input($c$.trainingError($x$)) \\
		\If{$dd$.driftState = $WARNING$}{$b$.add($x$)}
		\ElseIf{$dd$.driftState = $DRIFT$}
		{
			$C$.save($c$) \\
			$c_r \leftarrow C$.reuse($b$) \hfill \scriptsize{\emph{return copy of most accurate classifier on buffer}} \small \\
			$c_n \leftarrow$ new Classifier() \\
			$c_n$.trainOn($b$) \\
			$C$.representClassifiers($b$, $m$) \hfill \scriptsize{\emph{use buffer instances to find similar classifiers}} \small \\
			$C$.fadeClassifiers($f$) \hfill \scriptsize{\emph{fade classifiers not recently reused}} \small \\
			$b$.empty()
		}
		\Else{
			$c_r$.trainOn($x$) \\
			$c_n$.trainOn($x$) \\
			$b$.empty()
		}
	}
	\small
	\KwOut{Classification $h$}
	\caption{Pseudo-code for ECPF}
	\label{ECPFalg}
\end{algorithm}

ECPF uses two techniques to restrict the size of its classifier collection, using both techniques when drift occurs. When reusing a classifier, we may save a copy back into our classifier collection with the original and have two very similar classifiers. If the classifiers are no longer similar, we would prefer to keep both, assuming they now represent different concepts. If they are similar, we will keep the one with the best overall accuracy, and use it to \emph{represent} the other, which we delete. By using the more accurate classifier to represent the less accurate classifier, we protect ourselves from removing a very effective classifier in favour of a reasonably effective classifier. We discuss classifier similarity further below, as well as \emph{fading}, our other approach to constraining the size of the classifier collection.

ECPF trains two classifiers and so tends to be slower than CPF, but has a greater breadth of options for managing and reusing classifiers. It can switch between using the new or reused classifier based on accuracy since the last drift. CPF must decide whether to reuse a classifier based on a short buffer of instances after a drift, so requires a set buffer length $b$ to ensure that it can make an informed decision. ECPF can relax this requirement and adapt to a drift immediately, relying on the drift detector's warning period to select the classifier to reuse.

\subsection{Representing classifiers with similar classifiers in ECPF}

After drift is detected, ECPF will decide whether any classifiers can represent other classifiers that behave similarly. A user-set similarity parameter $m$ states what proportion of instances two classifiers should classify in the same way if they are to be considered as describing the same concept, where $0 \leq m \leq 1$. The longer two classifiers are in the classifier collection, the more instances will be seen in common between the two, with the law of large numbers suggesting that if two classifiers are similar, we will eventually be able to see this. Lower levels of $m$ may lead to erroneous matching of two classifiers which could lead to removing one which describes a different concept from the other.

Our approach uses a similarity measure based upon \emph{conceptual equivalence} used by \citet{gomes2012} for comparing classifiers. We adapt their approach and do pairwise comparisons of classifiers' respective errors when classifying given instances in warning buffers, when drift occurs. When comparing classifier $c_a$ and classifier $c_b$, we calculate a score per instance, where $c_a(x)$ and $c_b(x)$ is the classification error for $c_a$ and $c_b$ on a given instance $x$:

\[
Score(x, c_a, c_b) = 
\begin{cases}
1 & \text{if } c_a(x) = c_b(x) \\
0 & \text{if } c_a(x) \neq c_b(x)
\end{cases}
\]

We then calculate similarity for two classifiers over a range of instances, using instances $X$ of length $z$ seen during warning drift periods:

\[
Sim(X_1...X_z, c_a, c_b) = \frac{\Sigma^z_{i=1} (Score(X_i, c_a, c_b))}{z}
\]

If $Sim(X, c_a, c_b) \geq m$ where $m \leq 1$, we describe the classifiers as similar and likely to represent the same concept.  

\begin{figure}[h!]
	\begin{center}
		\includegraphics[width=\textwidth]{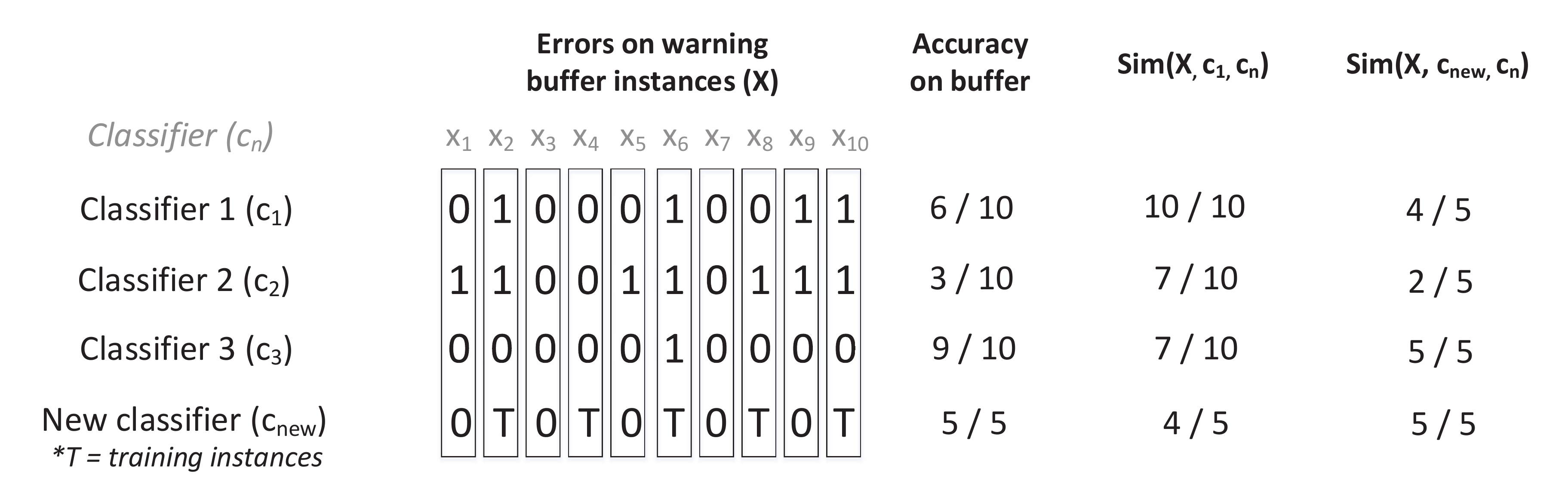}
	\end{center}
	\caption{Example of calculating classifier similarity for classifiers $c_1...c_{new}$ for instances in warning buffer}
	\label{Similarity}
\end{figure}

Every time a drift occurs, all existing classifiers will classify instances in the warning buffer and errors of each classifier will be compared. The similarity matrix stores pairwise comparisons between classifiers, through recording instances seen and total score (or $\Sigma (Score(X, c_a, c_b))$). We then check for similar classifiers. Figure \ref{Similarity} shows an example of four classifiers being compared using a given warning buffer, with 0 representing a correct classification and 1 an incorrect classification by the classifiers. Of ten instances $x_1 \dots x_{10}$ in our buffer, we can see which our classifiers correctly or incorrectly classify, which is the behaviour we measure to find their $Sim$. Using our first reuse approach, if $m = 0.95$, we can see that $Sim(X, c_{new}, c_3) = 5/5 \geq m$, so could treat these classifiers as similar, and keep one of them. In practice, we would require at least $b$ instances in the buffer to make that decision. We explore selecting $m$ in our experiments, but would generally recommend $0.85 \leq m \leq 0.99$. Lower levels of $m$ lead to ECPF have a smaller selection of more general classifiers, regularly representing classifiers with others. Higher levels will lead to a larger selection of more specific classifiers that may slow the framework but will provide better accuracy when many similar but distinct recurring concepts are present in a stream.  

\subsection{Fading}
\label{fadingSection}

Data streams can be of infinite length, and over time, our classifier collection may grow, risking continually increasing time and memory requirements. To avoid this, we use a fading mechanism to constrain the size of our classifier collection. Our fading mechanism ranks classifiers by their recency and how regularly they are reused. At any stage, an array of fade points $F$ shows how close each classifier is to being deleted. Fade points of a given classifier $c_a$, $F_{c_a}$ can be expressed as follows: 

\[
F_{c_a} = 
\begin{cases}
0 & \text{if represented by older classifier}\\
(r+1) \times f + \sum (F_{m_{c_a}}) - (d-r) & \text{otherwise}
\end{cases}
\]

Here, $r$ represents the number of drift points after which $c_a$ is reused, $f$ is a user-set parameter for points to gain on creation and reuse, $d$ is the number of drifts the classifier has existed for (excluding the drift at which it is created) and $\sum (F_{m_{c_a}})$ is the sum of points held by classifiers when we choose to represent them by $c_a$. Classifiers gain $f$ points when they are reused, and are penalised a point when they are not. When newer classifiers are represented by an older classifier, the older classifier receives all of the newer classifier's points. When $F_{c_a} = 0$, the classifier is deleted. The user can control the size and number of classifiers by selecting $f$. Fading is an optimisation that will generally reduce the number of stored classifiers when $f$ is set low, but cause increased memory use when $f$ is set high. With a higher value of $f$, ECPF can hold more classifiers and is more likely to have an appropriate classifier to reuse for a recurring concept when it does recur. This setting should always be set so it can accommodate the number of likely recurring concepts in the underlying data. For instance, if a shop has distinct customer behaviour for morning, noon and evening, then $f$ should be set to three at minimum. However, ECPF may hold multiple classifiers trained on the same concept that it has not yet identified as being of a similar concept. Therefore, it would make sense to set $f$ to more than three to allow ECPF more chance to compare classifiers to other before removing them.

This approach to controlling the size of the classifier collection could technically create an unbounded number of classifiers -- for example, if $f-1$ classifiers were repeatedly reused to gain a very large number of points, before a new set of distinct concepts emerged causing a different set of $f-1$ classifiers to gain large numbers of points, and so on. Where no classifiers are reused or represented by others, we cannot have more than $f$ classifiers at a given time. Fading points provide a ranking of classifiers based on recency and reuse. Instead of removing classifiers where they have zero points, the framework can remove the lowest ranked classifier when it uses a set amount of memory, which allows ECPF to utilise all available memory while functioning in a constrained environment. 

\subsection{Model management in Enhanced CPF}

\begin{figure}[]
	\begin{center}
		\includegraphics[width=\textwidth]{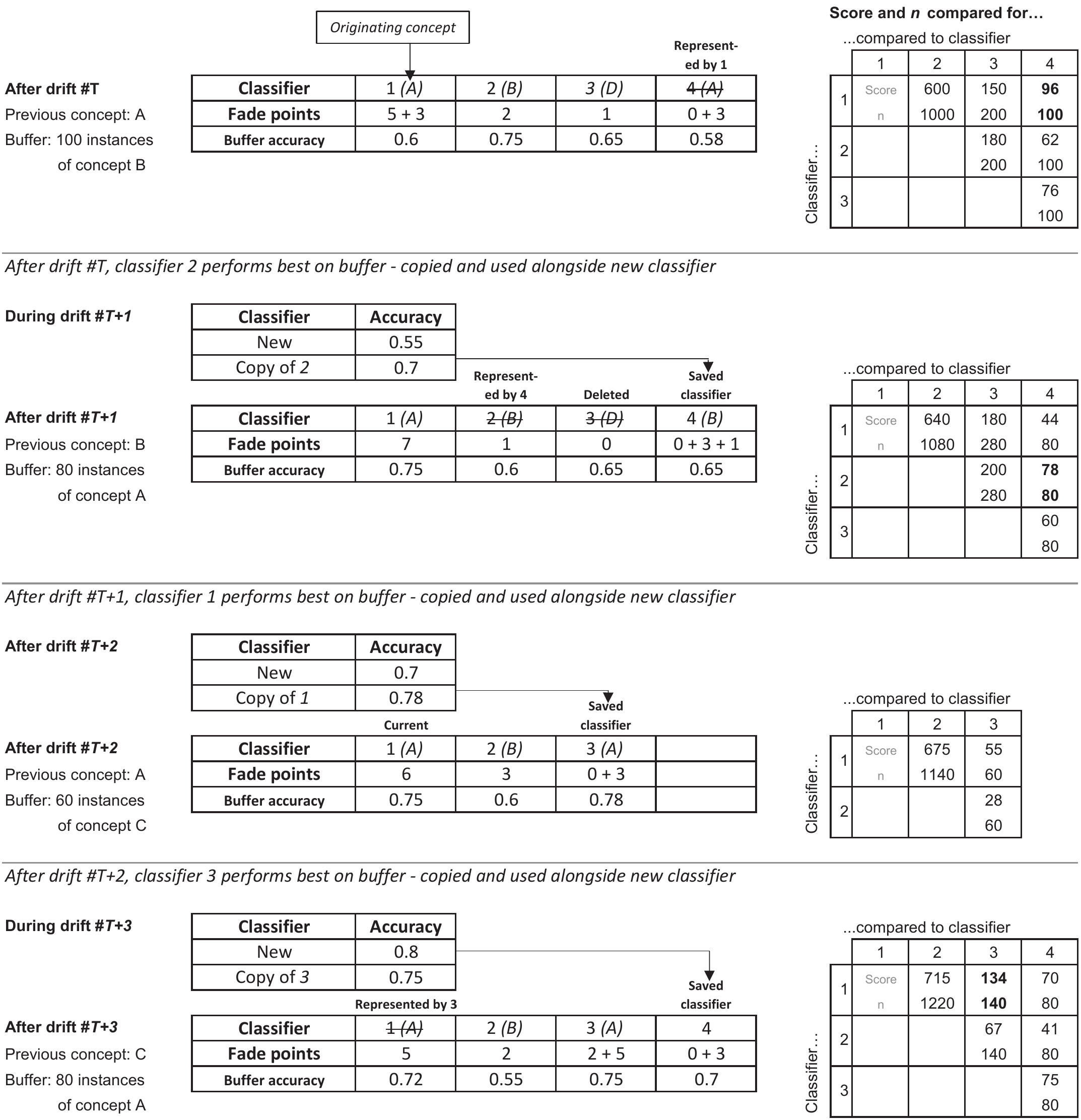}
	\end{center}
	\caption{Example of ECPF classifier management in practice ($n$ = comparison instances seen by two classifiers)}
	\label{ECPFclassifier}
\end{figure}

Fig. \ref{ECPFclassifier}  illustrates how ECPF maintains its collection of classifiers. We have set $f = 3$ and $m = 0.95$ for the example.

\begin{enumerate}
	\item After drift point $T$ in a stream, in which our warning buffer had 100 instances, we have three classifiers plus a copy of $c_1$, $c_4$ which was produced when $c_1$ was reused for the previous instances. The copy gains $f$ points, while $c_1$, $c_2$ and $c_3$ lost a point each. Pairwise comparisons in the similarity matrix have been updated with the $Score$ from 100 new instances seen in the warning buffer. This similarity made $c_4$ appear similar to classifier $c_1$ and since $c_1$ has better overall accuracy than $c_4$ (not shown), we keep $c_1$. $c_1$ therefore will gain the $f$ fade points that $c_4$ had. As $c_2$ worked best on the warning buffer, a copy will be produced to classify the following instances alongside a new classifier.
	
	\item The copy of $c_2$ outperforms the new classifier and is saved. After drift $T+1$, this classifier is saved as $c_4$ and gains $f$ points, while the others lose a point each. $c_2$ is now represented by $c_4$ due to having greater than $m$ similarity, and so $c_4$ gains $c_2$s fade point. $c_1$ achieves the best accuracy of existing classifiers on the warning buffer and is reused. Classifier $c_3$ now has zero points and is deleted, and $c_4$ becomes $c_3$.
	
	\item The copy of $c_1$ outperforms the new classifier and is saved. After drift $T+2$, $c_3$, the copy of $c_1$, gains $f$ points. The other two classifiers lose a point each. Even though $c_3$ was a copy of $c_1$, after being trained on instances after drift $T+1$, it is not conceptually equivalent, so we keep $c_1$ and $c_3$. A new concept, C appears, and $c_3$ performs best on the warning buffer, so gets reused.
	
	\item The copy of $c_3$ is outperformed by the new classifier, so the new classifier is saved. After classifying the next set of warning buffer instances, $c_1$ now looks conceptually equivalent to $c_3$ and as $c_3$ has a better overall accuracy (not shown), so $c_1$ is represented by $c_3$.
\end{enumerate}

\subsection{Theoretical discussion of ECPF}

In data streams, techniques that have super-linear time and memory complexity in regards to instances seen will become unusable over many instances. Provided classifiers and the drift detector used in the framework have linear or better time and memory complexity, ECPF will as well. In respect to time, no more than two classifiers are ever trained on a given instance. All classifiers held are tested on warning buffer instances; the size of the classifier collection is constrained in two ways. The fading parameter $f$ will in almost all cases limit the number of classifiers that can be held in the classifier collection. Across all of our experiments, the classifier collection never held more than $2f$ classifiers. A user-set memory usage limit will lead to removal of classifiers with the least fade points when met. In regards to memory, ECPF's memory usage is governed by the size of the classifier collection and its classifier similarity matrix which records similarity of classifier behaviour on warning buffer instances. These are both capped, based on the size of the classifier collection. Finally, the instance buffer stores instances seen during warning periods and should be limited in size by the drift detector.

In a best case scenario, ECPF identifies two recurring concepts, $A$ and $B$, and reliably reuses a given classifier for each concept. When concept $A$ occurs the first time, ECPF trains a new classifier for it, and when concept B occurs, ECPF stores the more accurate of a new classifier and the reused concept A classifier. When concept A recurs, ECPF reuses the original concept $A$ classifier, creating a copy to train on new data. This provides better accuracy than a new classifier, and when saving this classifier back to the collection, it matches the behaviour of the original concept A classifier, so represents the original. When concept $B$ recurs, ECPF reuses the original concept $B$ classifier, and at the next drift, it saves back the better trained copy which will represent the original. In this way, ECPF stores no more than two classifiers in its collection while training a copy of an existing classifier and a new classifier at any given time.

In a worst case scenario, no concepts recur. ECPF reuses an inappropriate classifier whenever drift occurs, which is less accurate than the new classifier, which is stored. The classifier collection is filled with distinct classifiers, none of which can be used to represent another. In this case, classifiers are not regularly reused and the fading parameter deletes classifiers as they reach zero fading points. This leads to a consistent maximum size and similar classification accuracy to a simple drift detection framework. In the unlikely event that existing classifiers are reused in a uniform fashion, the memory limit will lead to those with fewest fade points being deleted, which will lead to a consistent memory size at the user set limit.

\section{Experimental Design and Results}

In this section, we detail the experiments we have used to validate the performance of our proposed techniques. We divide our experiments into three sections. The first compares the \textbf{performance} of ECPF against state-of-the-art and recent ensembles and frameworks that reuse classifiers. We show accuracy, Kappa, runtime and memory usage across real-world and synthetic datasets, to show that our proposed technique is reliably faster while providing at least comparable accuracy to state-of-the-art approaches. In our next set of experiments, we test the \textbf{robustness} of ECPF compared to state-of-the-art techniques, testing against synthetic datasets with multiple classes, attribute and class noise, shorter periods of drifts, class imbalance and different classifiers. We show ECPF is not unduly affected by these characteristics of data streams. Finally, we test ECPF with varied drift detectors and parameters to show how these choices affect its \textbf{behaviour}. For our experiments, we used the MOA API (available from \url{http://moa.cms.waikato.ac.nz}). Experiments were run on an Intel Core i5-4670 3.40 GHz Windows 7 system with 8 GB of RAM.

First, we list the metrics used for comparing techniques across our experiments. For runtime, we measure runtime (in milliseconds) for an entire framework (though not data generation). For memory use, we measure the size of all components of the framework. To measure the performance of our classifier predictions, we show prequential accuracy (\emph{i.e} based on a test-and-train approach). This is the proportion of correctly classified instances in the data stream across all instances. Where we wish to see how well a classifier is classifying with respect to minority classes, we also show the Kappa measure \citep{bif2010}. We show this measure in place of accuracy where balance of accurate classifications is of interest. 

In our experiments, we characterise ECPF by how it is affected by different drift detectors. A perfect drift detector would detect a concept drift every occasion that the underlying concept changes in a stream. However, in practice, most drift detectors will either underestimate or overestimate the drifts in the underlying stream, depending on the nature of the problem and the signals they are relying on to detect drift. In our experiments testing ECPF and CPF with real drift detectors on synthetic data, we show the number of drifts detected and compare this to the number of true concept drifts that have occurred to understand whether the drift detector combined with our classification framework has been oversensitive or undersensitive in detecting drift.

\subsubsection{Comparison approaches}

We have used Diversity Pool (DP) \citep{chi2018} and Adaptive Random Forests (ARF) \citep{gom2017} as baseline techniques to compare against. These are state-of-the-art classifier reuse and ensemble techniques respectively. We also tested a recent ensemble technique AUE \citep{brz2014} and classifier reuse framework RCD \citep{goncalves2013} in experiments testing the performance of our framework. Through all of our experiments, we also 
contrast ECPF with the original CPF framework. For ECPF, we set our classifier fading parameter $f=15$, our similarity measure $m=0.95$ and for CPF, our minimum buffer size as $b=60$ unless otherwise stated. Implementations of our algorithms can be found at \url{https://github.com/rand079/CPF}. Default parameters as per MOA implementations are used for ARF, AUE and DP. 

Our framework allows the use of any incremental streaming classifier. Through our experiments, we use Hoeffding Trees with Na\"\i ve Bayes. This is a version of a CVFDT \citep{hul2001} which creates a Na\"\i ve Bayes classifier at the leaves if it provides better accuracy than using the majority class (specifically \\\texttt{HoeffdingTree} with \texttt{leafPredictionOption = 1} in MOA). This is consistent with the implementation of Diversity Pool made available by the authors, though distinct from the classifier type used in \citet{and2016}. We used this classifier type in all comparison techniques, except for Adaptive Random Forests which uses its own specifically designed Hoeffding Tree.

In experiments with synthetic data, instead of using an actual drift detector, we signal a warning to our techniques when actual drift occurs, and a drift sixty instances afterwards. This emulates drift detector behaviour while not affecting our experiments based on the characteristic behaviour of the detector. We also test actual drift detectors with synthetic data. We test HDDM-A and RDDM as the leading drift detectors as per \cite{bar2018}, using default parameters from MOA implementations. We include MagSeed as a version of ADWIN with an added warning bound, and use the author's recommended parameters: $\delta = 0.05$, $\epsilon = 0.01$, $\alpha = 0.8$ and $\delta_w = 0.1$. CPF, DP and RCD always use the same drift detector as ECPF in our experiments.

\subsubsection{Datasets}

We test comparison techniques across data created by five synthetic dataset generators thirty times, generating different synthetic data each time. Each stream has 400 abrupt drift points with 2500 instances between each drift point. Concepts recur sequentially for all of our datasets. All data stream generators are available in MOA apart from CIRCLES, which uses a linear circular class separator with centre point (0.5, 0.5) and a changing radius. All dataset generators not included in MOA are available from \url{https://github.com/rand079/CPF}.

\begin{table}[htbp]
	\centering
	\caption{Synthetic data used for experiments}
	\resizebox{0.92\textwidth}{!}{
		\begin{tabular}{ccccccc}
			\toprule
			\textbf{Dataset} & \textbf{Classes} & \textbf{Attributes} & \textbf{Att types} & \textbf{Concepts} & \textbf{Concept variable} & \textbf{Concept values} \\
			\midrule
			\textbf{Agrawal} & 2     & 9     & Numeric & 5     & function & {1, 3, 5, 7, 9} \\
			\midrule
			\textbf{CIRCLES} & 2     & 2     & Numeric & 5     & radius & {0.2, 0.25, 0.3, 0.35, 0.4} \\
			\midrule
			\textbf{LED} & 10    & 10    & Binary & 4     & numberAttributesDrift & {1, 3, 5, 7} \\
			\midrule
			\textbf{RandomRBF} & 2     & 10    & Numeric & 5     & numCentroids & {10,20,30,40,50} \\
			\midrule
			\textbf{STAGGER} & 2     & 3     & String & 3     & function & {1, 2, 3} \\
			\bottomrule
		\end{tabular}%
	}
	\label{synthDatasets}%
\end{table}%

 We use eight real-world datasets, commonly used for benchmarking data stream algorithms, to compare our approaches: \fnurl{Electricity, Airlines, Poker Hand and Covertype}{http://moa.cms.waikato.ac.nz/datasets}; \fnurl{Network Intrusion}{http://kdd.ics.uci.edu/databases/kddcup99/kddcup99.html}; \fnurl{Sensor }{http://www.cse.fau.edu/~xqzhu/Stream/sensor.arff} \citep{zhu2011}; \fnurl{Social Media}{https://archive.ics.uci.edu/ml/datasets/Buzz+in+social+media+} \citep{kaw2013}; and \fnurl{Power Supply }{http://www.cse.fau.edu/~xqzhu/Stream/powersupply.arff} \citep{UCRArchive}.
 
\begin{table}[htbp]
	\caption{Real-world datasets used for experiments}
	\centering
	\resizebox{0.6\textwidth}{!}{
		\begin{tabular}{ccccc}
			\toprule
			\textbf{Dataset} & \textbf{Classes} & \textbf{Attributes} & \textbf{Att types} & \textbf{Instances} \\
			\midrule
			\textbf{Electricity} & 2     & 8     & Numeric & 45,312 \\
			\midrule
			\textbf{Airlines} & 2     & 7     & Numeric & 539,383 \\
			\midrule
			\textbf{Intrusion} & 23    & 42    & Catg./Numeric & 494,021 \\
			\midrule
			\textbf{Covtype} & 10    & 54    & Binary & 581,012 \\
			\midrule
			\textbf{Poker} & 10    & 10    & Numeric & 829,012 \\
			\midrule
			\textbf{Sensor} & 54    & 5     & Numeric & 2,219,803 \\
			\midrule
			\textbf{Power Supply} & 24    & 2     & Numeric & 29,928 \\
			\midrule
			\textbf{Social Media} & 2     & 77    & Numeric & 140,707 \\
			\bottomrule
		\end{tabular}%
	}
	\label{realDatasets}%
\end{table}%

\subsection{Experiments testing the performance of ECPF}

In this section, we provide a comparison of ECPF against recent and state-of-the-art approaches when classifying data streams. We show that our proposed algorithm is faster than comparison approaches apart from CPF while competitively accurate across a variety of synthetic and real-world datasets.



For these experiments, we classified data on streams from our five synthetic dataset generators and eight real-world data streams regularly used for testing stream classification performance. On graphs, we show a 95\% confidence interval as error bars, assuming normality of results. These are included for all plots, but in some cases, variance of results is so low that they cannot be observed. We test ECPF against DP and ARF as state-of-the-art stream classifiers and RCD and AUE as recent techniques. All are described in Section 2. We contrast classification accuracy and Kappa with runtime and memory usage. For accuracy and Kappa, higher values are considered better, while for runtime and memory, lower values are better. We perform post-hoc Nemenyi testing ($\alpha = 0.95$) to measure statistically significant differences in ranking across techniques for each measure. These plots are ranked from highest to lowest, so high rankings are preferable for accuracy and Kappa but not  for runtime and memory use.

\begin{figure}[!htbp]
	\begin{minipage}[t]{0.45\textwidth}
		\centering
		\includegraphics[width=0.9\textwidth]{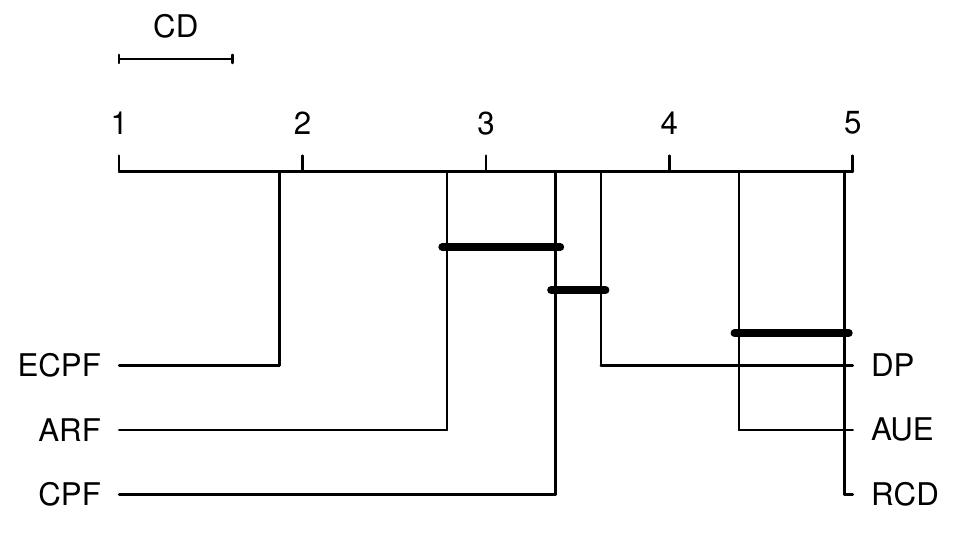}
		\vspace{-0.5em}
		\caption*{Accuracy}
		\label{plotSynthAccNemenyi}
	\end{minipage}
	\hfill
	\begin{minipage}[t]{0.45\textwidth}
		\centering
		\includegraphics[width=0.9\textwidth]{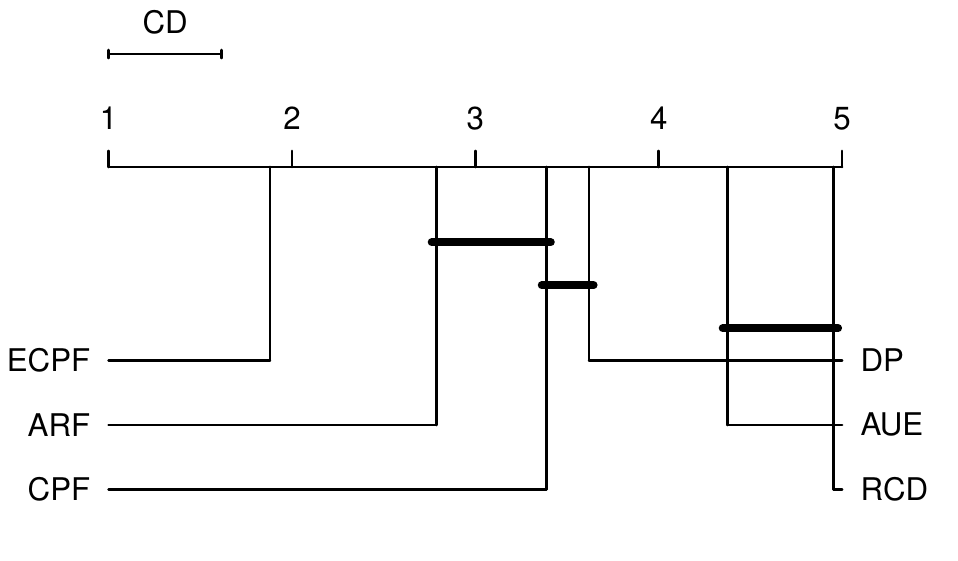}
		\vspace{-0.5em}
		\caption*{Kappa}
		\label{plotSynthKappaNemenyi}
	\end{minipage}
	\hfill
	\begin{minipage}[t]{0.45\textwidth}
		\centering
		\includegraphics[width=0.9\textwidth]{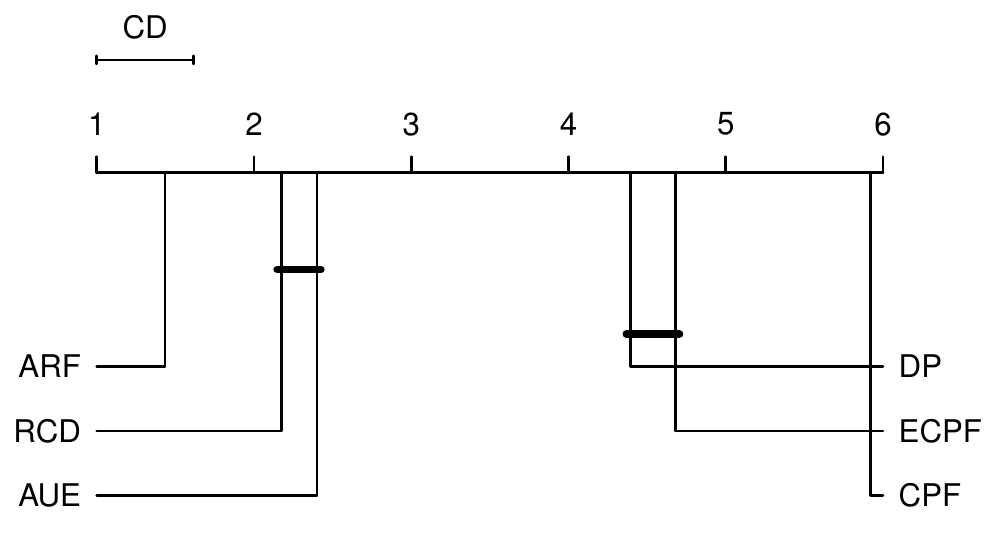}
		\vspace{-0.5em}
		\caption*{Runtime}
		\label{plotSynthMemNemenyi}
	\end{minipage}
	\hfill
	\begin{minipage}[t]{0.45\textwidth}
		\centering
		\includegraphics[width=0.9\textwidth]{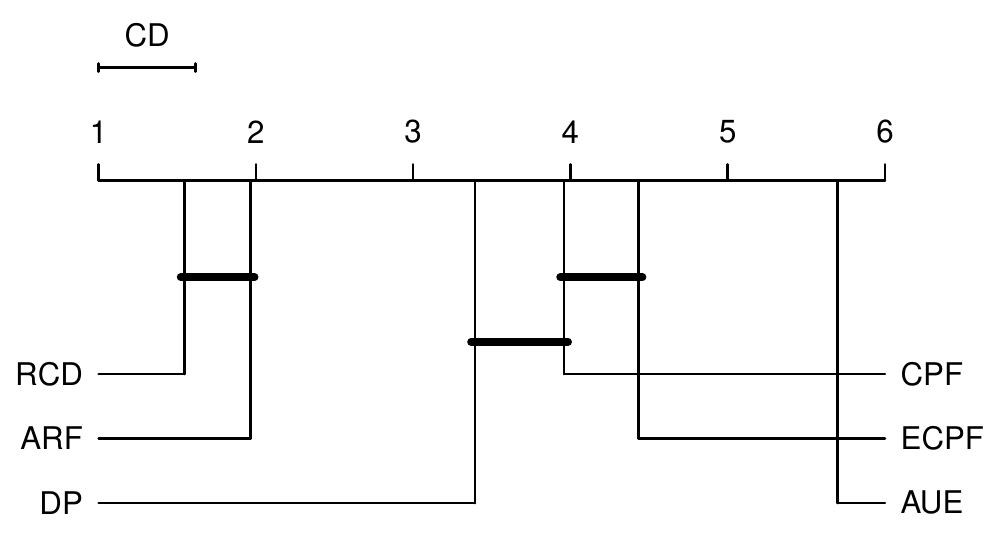}
		\vspace{-0.5em}
		\caption*{Memory use}
		\label{plotSynthTimeNemenyi}
	\end{minipage}
	
	\caption{Critical difference plots from posthoc Nemenyi tests ($\alpha = 0.95$) of rankings for experiments on synthetic datasets. Rankings left-to-right respectively represent largest-to-smallest values.}
	\label{plotNemenyiSynth}
\end{figure}

Fig. \ref{tabSynth} show results of our experiments on synthetic datasets, with Fig. \ref{plotNemenyiSynth} showing critical difference plots of each measure. ECPF was statistically the highest ranked in terms of accuracy and Kappa measure across datasets. As these datasets have regularly recurring concepts, they are ideal for frameworks with classifier reuse. ARF, RCD and AUE were statistically significantly slower than ECPF. Though ECPF was not statistically significantly faster than DP, it ran faster on all datasets but for RandomRBF. ECPF used significantly less memory than RCD and ARF, and more than AUE. CPF ran significantly faster than ECPF, but required more memory. CPF retains classifiers that have been trained on more data over those trained on less. This is likely to lead to more complex trees being held within its classifier collection.


\begin{table}
	\scriptsize
	\caption{Accuracy, Kappa, time and memory usage across synthetic datasets (best results in bold)}
	\label{tabSynth}
\begin{tabular}{p{1.25cm}ccccccccc}
	\toprule
	Stream & Framework & Acc & $\sigma$ & Kappa & $\sigma$ & \shortstack{Time \\ (s)} & $\sigma$ &  \shortstack{Memory \\ (000 bytes)} & $\sigma$\\
	\midrule
	Agrawal & ECPF & \textbf{82.2\%} & 1.1\% &  \textbf{64.5\%} & 2.2\% & 5.7 & 0.2 & 1010 & 440\\
	Agrawal & CPF & 78.2\% & 0.4\% & 56.4\% & 0.8\% & \textbf{3.6} & 0.1 & 490 & 100\\
	Agrawal & AUE & 74.4\% & 0.1\% & 48.6\% & 0.2\% & 40.4 & 1.0 &  \textbf{180} & 30\\
	Agrawal & DP & 65.3\% & 2.6\% & 30.6\% & 5.1\% & 6.6 & 3.1 & 5310 & 2330\\
	Agrawal & RCD & 65.4\% & 2.2\% & 30.7\% & 4.4\% & 34.6 & 4.1 & 6570 & 770\\
	Agrawal & ARF & 69.9\% & 0.3\% & 39.5\% & 0.6\% & 226.7 & 1.8 & 11870 & 4890\\
	\hline
	\addlinespace
	CIRCLES & ECPF & 93.4\% & 0.1\% & 86.8\% & 0.2\% & 2.1 & 0.0 & 190 & 30\\
	CIRCLES & CPF & 93.3\% & 0.3\% & 86.6\% & 0.6\% &  \textbf{1.4} & 0.0 & 370 & 40\\
	CIRCLES & AUE & 94.3\% & 0.0\% & 88.6\% & 0.1\% & 14.0 & 0.4 &  \textbf{110} & 10\\
	CIRCLES & DP & 92.5\% & 1.1\% & 84.9\% & 2.2\% & 4.0 & 1.7 & 310 & 180\\
	CIRCLES & RCD & 90.8\% & 0.2\% & 81.5\% & 0.4\% & 33.3 & 4.3 & 720 & 70\\
	CIRCLES & ARF & \textbf{94.9\%} & 0.0\% &  \textbf{89.7\%} & 0.1\% & 60.2 & 0.8 & 710 & 80\\
	\hline
	\addlinespace
	LED & ECPF & \textbf{72.1\%} & 0.1\% &  \textbf{69.0\%} & 0.1\% & 10.6 & 0.2 & \textbf{270} & 50\\
	LED & CPF & 70.8\% & 0.1\% & 67.5\% & 0.1\% & \textbf{7.5} & 0.1 & 440 & 0\\
	LED & AUE & 61.0\% & 0.1\% & 56.7\% & 0.1\% & 99.3 & 3.0 & 340 & 20\\
	LED & DP & 64.3\% & 1.4\% & 60.4\% & 1.6\% & 28.3 & 8.5 & 510 & 220\\
	LED & RCD & 65.4\% & 1.3\% & 61.6\% & 1.4\% & 57.3 & 22.4 & 2030 & 150\\
	LED & ARF & 61.8\% & 0.1\% & 57.6\% & 0.2\% & 82.1 & 0.9 & 640 & 100\\
	\hline
	\addlinespace
	RandomRBF & ECPF & 83.1\% & 1.7\% & 65.8\% & 3.3\% & 6.7 & 0.1 & 640 & 270\\
	RandomRBF & CPF & 78.1\% & 2.7\% & 55.6\% & 5.3\% &  \textbf{4.3} & 0.1 & 520 & 140\\
	RandomRBF & AUE & 75.6\% & 1.9\% & 50.5\% & 3.5\% & 52.0 & 1.1 &  \textbf{300} & 430\\
	RandomRBF & DP & 83.7\% & 1.9\% & 67.1\% & 3.9\% & 6.2 & 2.6 & 2240 & 370\\
	RandomRBF & RCD & 79.6\% & 1.5\% & 58.6\% & 2.8\% & 252.4 & 24.4 & 2020 & 170\\
	RandomRBF & ARF &  \textbf{89.4\%} & 1.0\% & \textbf{78.5\%} & 1.9\% & 128.2 & 4.5 & 1890 & 2320\\
	\hline
	\addlinespace
	STAGGER & ECPF &  \textbf{99.7\%} & 0.0\% &  \textbf{99.4\%} & 0.0\% & 1.3 & 0.1 & 180 & 20\\
	STAGGER & CPF & 98.3\% & 0.0\% & 96.6\% & 0.0\% &  \textbf{0.7} & 0.0 & 180 & 0\\
	STAGGER & AUE & 85.2\% & 0.0\% & 69.9\% & 0.1\% & 8.6 & 0.3 &  \textbf{100} & 10\\
	STAGGER & DP & 99.5\% & 0.0\% & 99.0\% & 0.1\% & 2.4 & 0.1 & 150 & 10\\
	STAGGER & RCD & 79.0\% & 0.5\% & 56.6\% & 0.9\% & 9.6 & 1.0 & 440 & 30\\
	STAGGER & ARF & 99.0\% & 0.0\% & 98.0\% & 0.0\% & 23.5 & 0.2 & 320 & 10\\
	\bottomrule
\end{tabular}
\end{table}

\begin{figure}[!htbp]
	\centering
	\includegraphics[width=\textwidth]{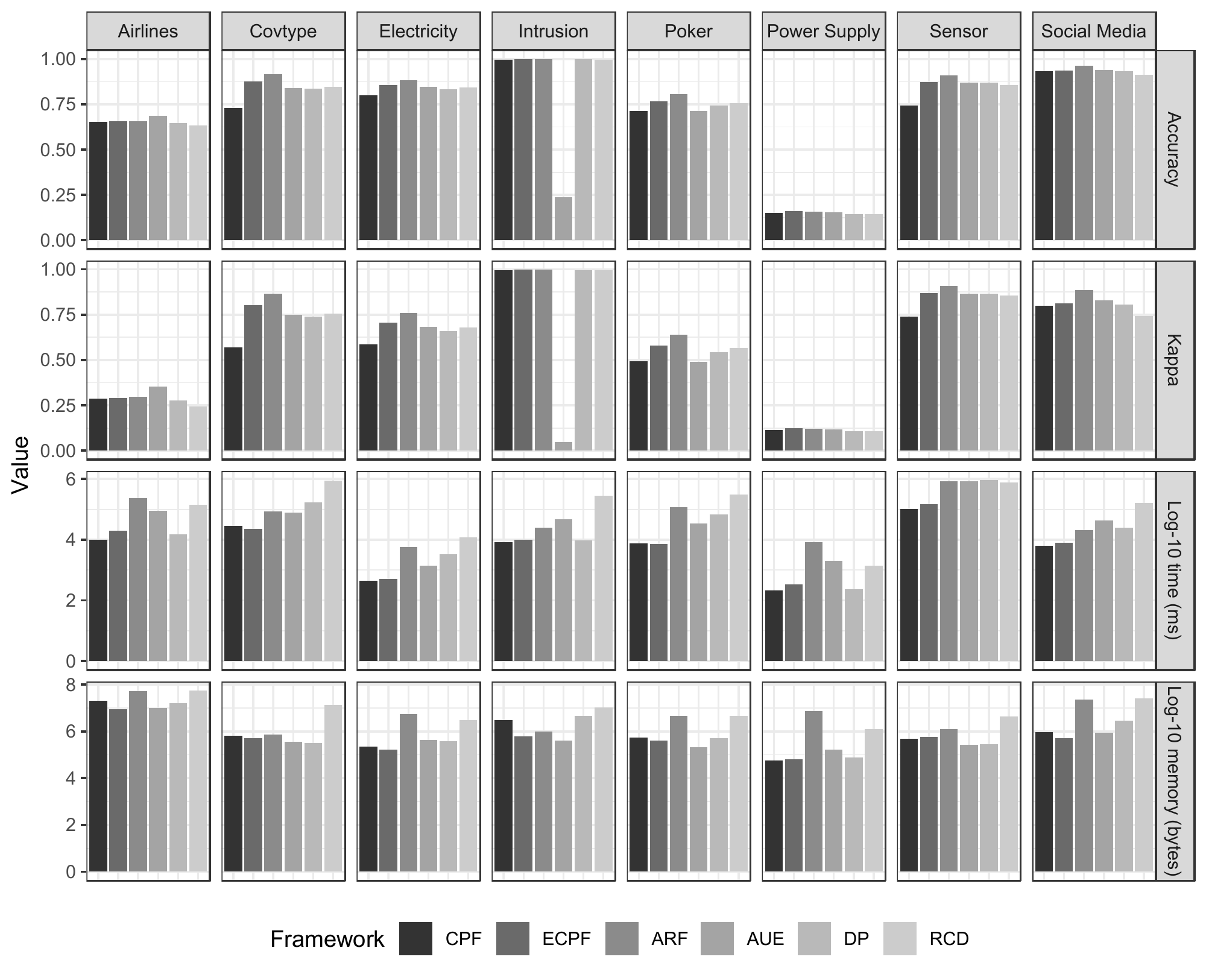}
	\caption{Accuracy, Kappa, time and memory usage across real-world datasets}
	\label{plotReal}
\end{figure}

\begin{figure}[!htbp]
	\begin{minipage}[t]{0.45\textwidth}
		\centering
		\includegraphics[width=0.9\textwidth]{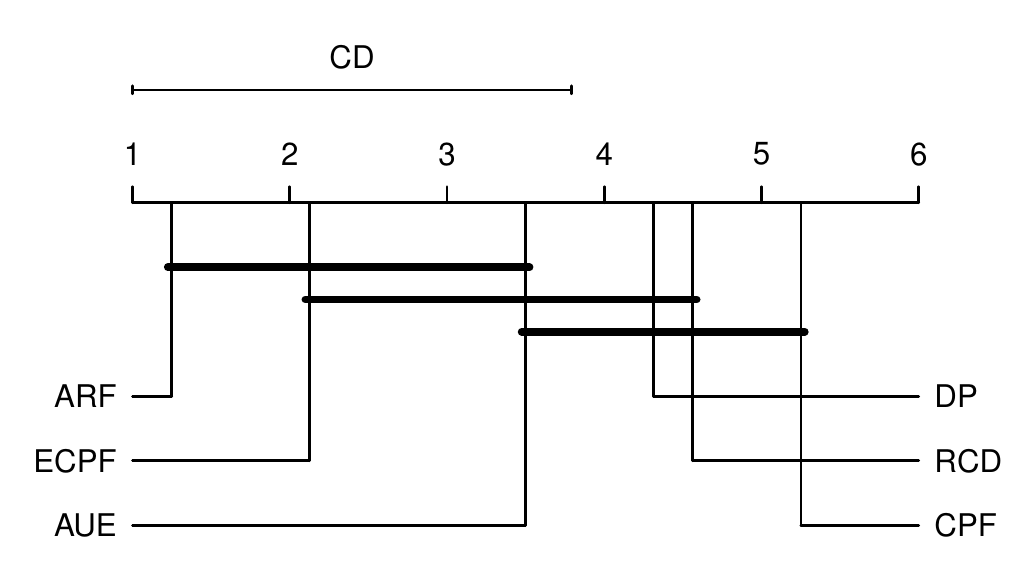}
		\vspace{-0.5em}
		\caption*{Accuracy}
		\label{plotAccNemenyi}
	\end{minipage}
	\hfill
	\begin{minipage}[t]{0.45\textwidth}
		\centering
		\includegraphics[width=0.9\textwidth]{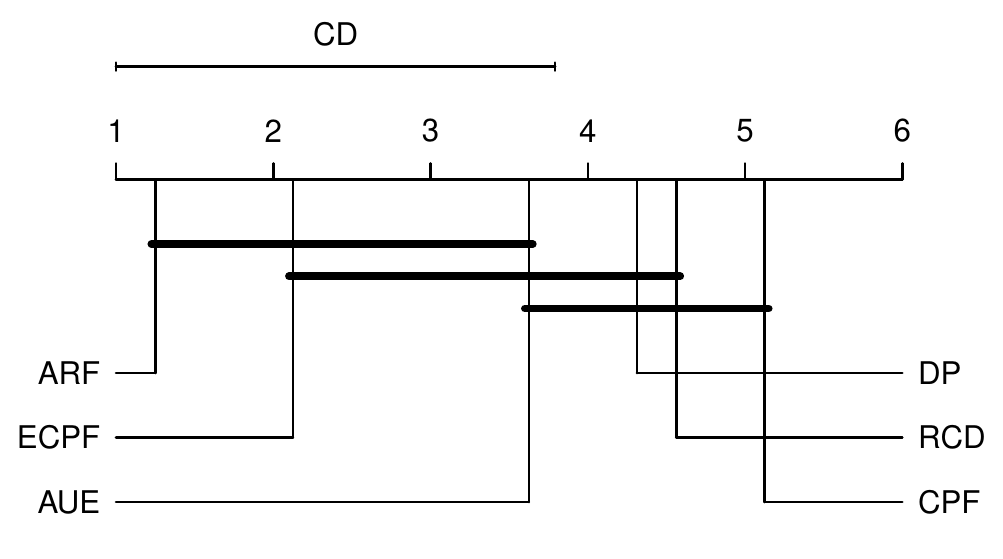}
		\vspace{-0.5em}
		\caption*{Kappa}
		\label{plotKappaNemenyi}
	\end{minipage}
	\hfill
	\begin{minipage}[t]{0.45\textwidth}
		\centering
		\includegraphics[width=0.9\textwidth]{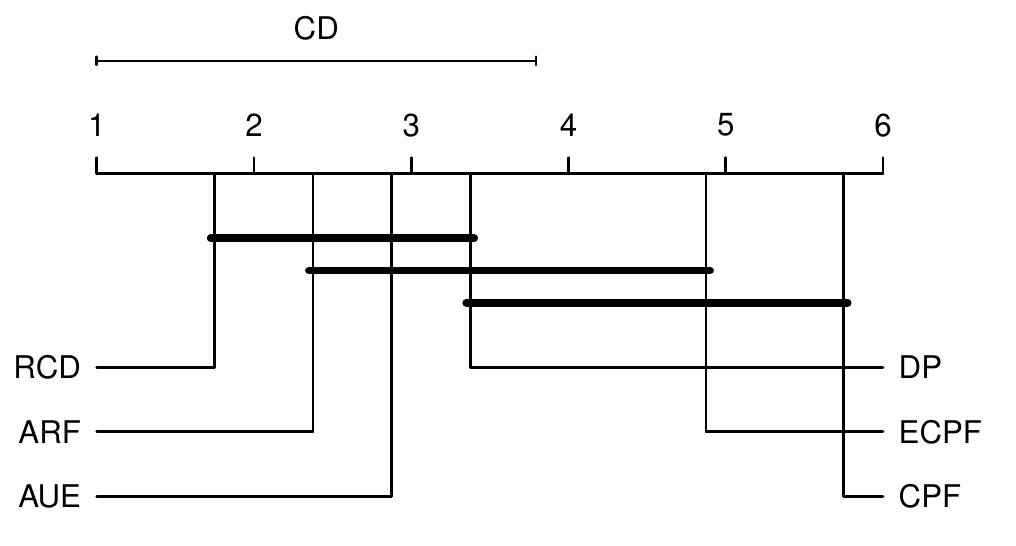}
		\vspace{-0.5em}
		\caption*{Runtime}
		\label{plotTimeNemenyi}
	\end{minipage}
	\hfill
	\begin{minipage}[t]{0.45\textwidth}
		\centering
		\includegraphics[width=0.9\textwidth]{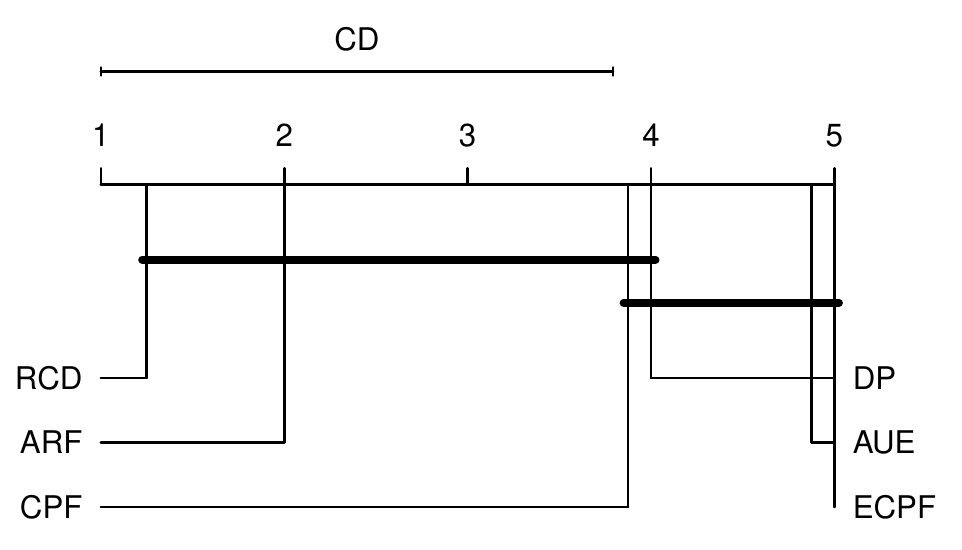}
		\vspace{-0.5em}
		\caption*{Memory use}
		\label{plotMemNemenyi}
	\end{minipage}
	\caption{Critical difference plots from posthoc Nemenyi tests ($\alpha = 0.95$) of rankings for experiments on real-world datasets. Rankings left-to-right respectively represent largest-to-smallest values.}
	\label{plotNemenyiReal}	
\end{figure}

Fig. \ref{plotReal} show results of our experiments on real-world datasets, with Fig. \ref{plotNemenyiReal} showing critical difference plots. These datasets did not reliably contain recurring concepts, and may have had more gradual concept drifts between concepts which could be expected to better suit ensembles than classifier reuse frameworks. On these datasets, ARF was generally the most accurate, with ECPF the second most. ECPF achieved within $4\%$ of ARF's accuracy across all datasets. DP was never as accurate as ECPF. CPF was significantly less accurate than ARF and ECPF, with notably poorer performance on CovType and Sensor datasets. ECPF is able to use classifications from a new classifier where it has chosen a poor classifier to reuse, but CPF cannot do so, so is penalised when it poorly reuses classifiers. Kappa followed a similar trend to accuracy, showing no techniques to be reliant on majority classes to gain high accuracy. CPF and ECPF were the fastest techniques amongst those tested, and though DP was not seen as being significantly faster, it was notably slower on many datasets, such as Covtype and Poker. ECPF's combined time to classify all real-world data streams was under a fifth of DP's. DP was never faster than CPF, and was only marginally faster than ECPF on three datasets, Airlines, Intrusion and Power Supply. CPF and ECPF required significantly less memory than ARF and RCD but not DP.

\subsection{Experiments testing the robustness of ECPF and CPF}

ECPF relies on warning periods before drift detection to select classifiers to reuse, based upon accuracy or similarity between a new classifier and existing classifiers in our classifier set. The experiments in these sections show that this approach is robust across different classifier types and dataset characteristics. We show that our techniques maintain performance against the state-of-the-art comparison techniques (ARF and DP) regardless of: the period between drifts; the class balance of the underlying dataset; in the presence of noise in attribute variables; and in the presence of noise in the class variable. We also show that ECPF maintains its advantage over DP when using different classifier types. Combined, these experiments show ECPF provides performance at least as robust as slower, more complex techniques.

\begin{figure}[!htbp]
	\centering
	\includegraphics[width=\textwidth]{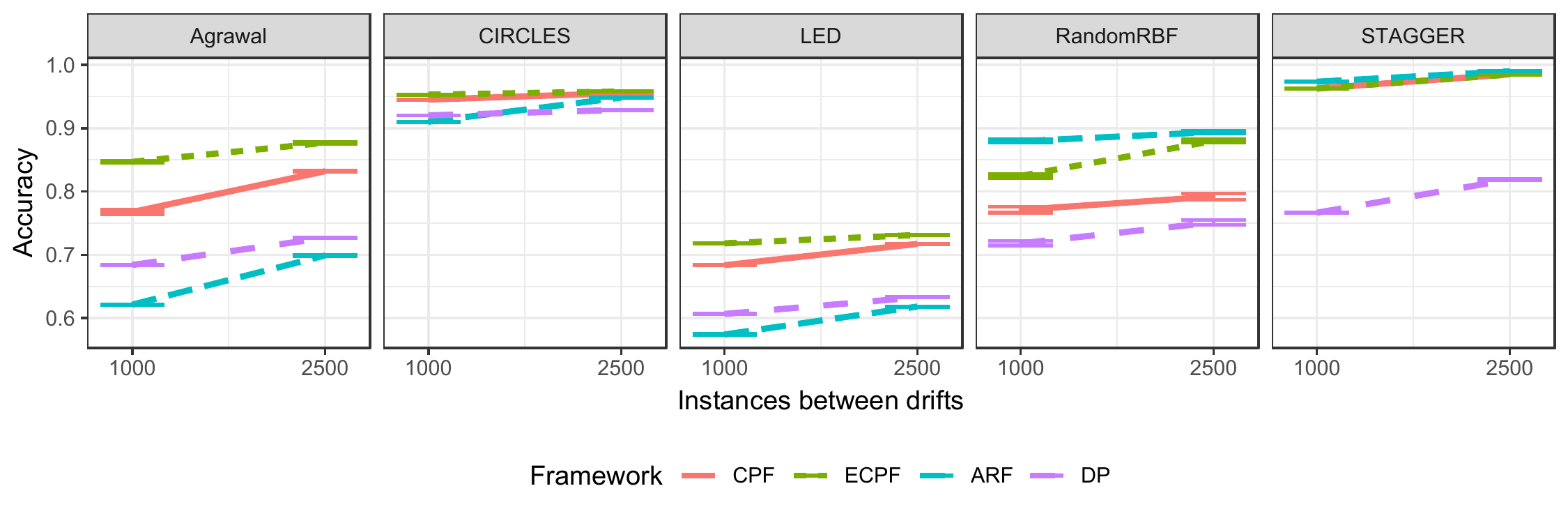}
	\caption{Accuracy on synthetic datasets with drifts every 1000 and 2500 instances}
	\label{plotSynthPeriod}
\end{figure}

In Fig. \ref{plotSynthPeriod}, we show accuracy for CPF, ECPF, ARF and DP on datasets with drifts every 1000 instances and every 2500 instances, with the same number of total drifts. Shorter periods between drifts can lead to less time to build developed classifiers. This could penalise classifier reuse approaches, as classifiers may not perform particularly well when the concept they were trained on recurs. However, in practice, the shorter drift period affected ARF the most in Agrawal, CIRCLES and LED, ECPF in RandomRBF and DP in STAGGER. Our techniques appear to classify relatively well in the presence of short drift periods.

\begin{figure}[!htbp]
	\centering
	\includegraphics[width=\textwidth]{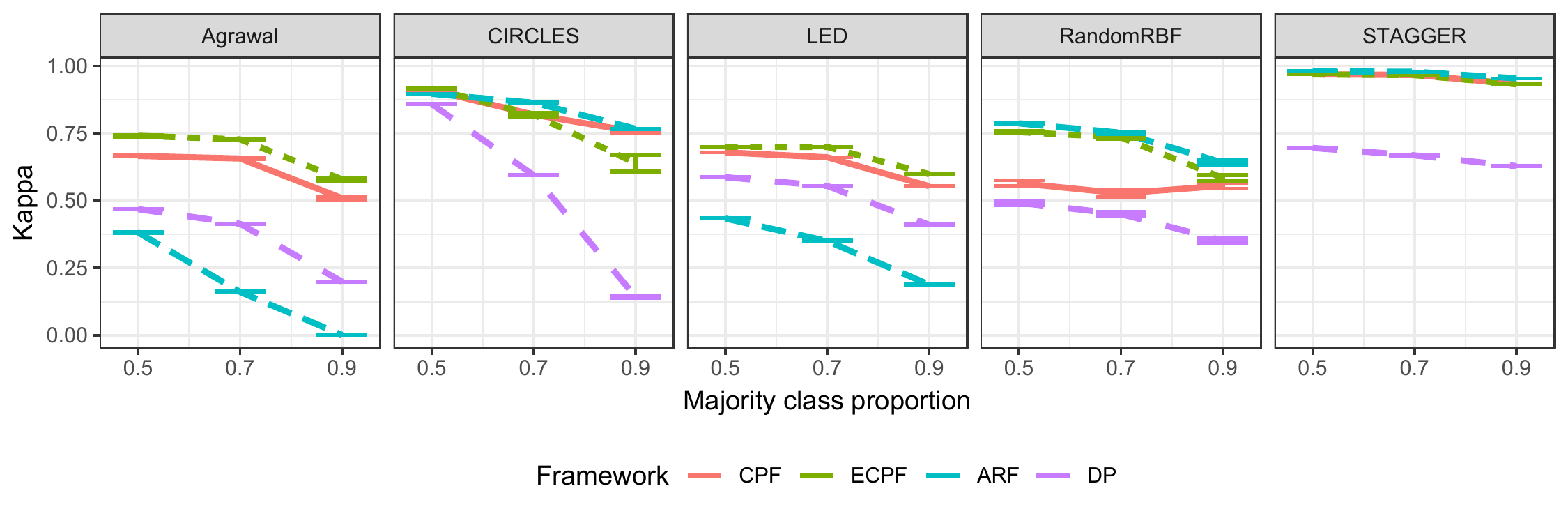}
	\caption{Kappa on synthetic datasets with $0.5, 0.7, 0.9$ instances proportionately of the majority class}
	\label{plotSynthBal}
\end{figure}

In Fig. \ref{plotSynthBal}, we show Kappa for CPF, ECPF, ARF and DP on datasets with varied class balance by changing how many instances of the majority class are generated. As our techniques rely on using classification similarity to identify classifiers to reuse, a change in class balance may make it more difficult to recognise appropriate classifiers to reuse, as there may not be a reasonable variety of classes being compared. When the Kappa measure is close to zero, it suggests that a learning framework does not clearly outperform classifying randomly by the class balance. All techniques achieved the best Kappa with half of the instances being of the majority class (all datasets are binary except for LED which has ten classes). Our experiments show that ARF dropped most in terms of Kappa with greater class imbalance on Agrawal and LED, DP dropped most on CIRCLES and STAGGER and ECPF dropped most on RandomRBF. Our techniques appear to classify relatively well in the presence of class imbalance.

\begin{figure}[!htbp]
	\centering
	\includegraphics[width=\textwidth]{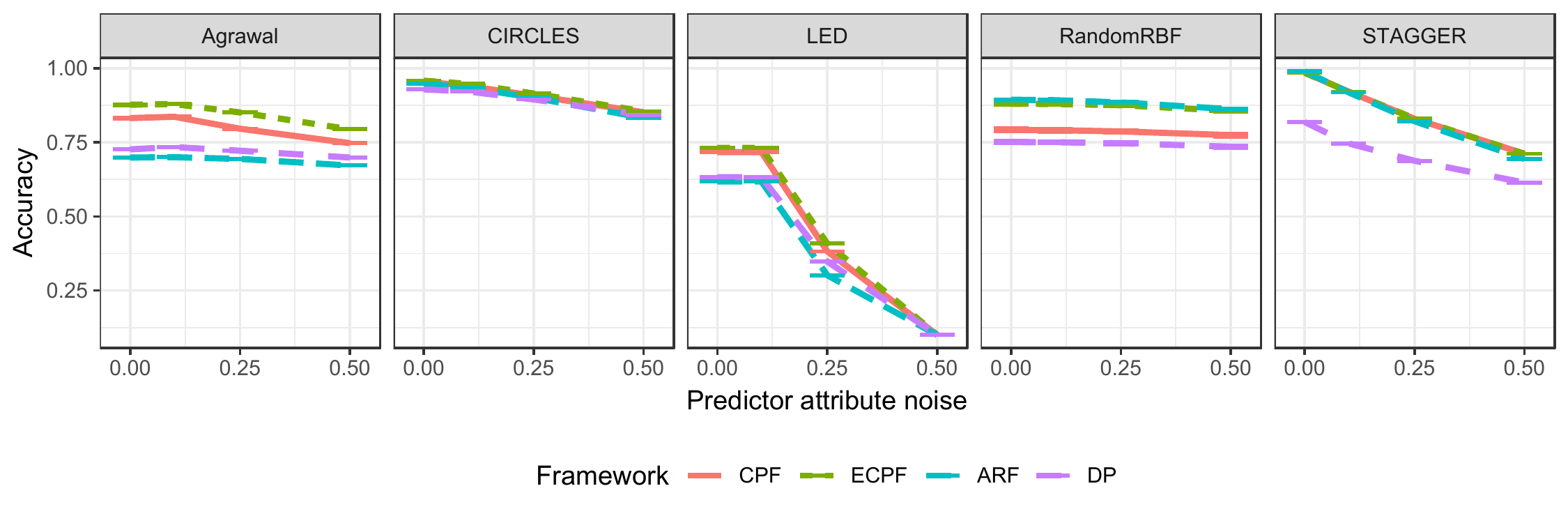}
	\caption{Accuracy on synthetic datasets with noise applied to predictor attributes}
	\label{plotSynthNoise}
	\includegraphics[width=\textwidth]{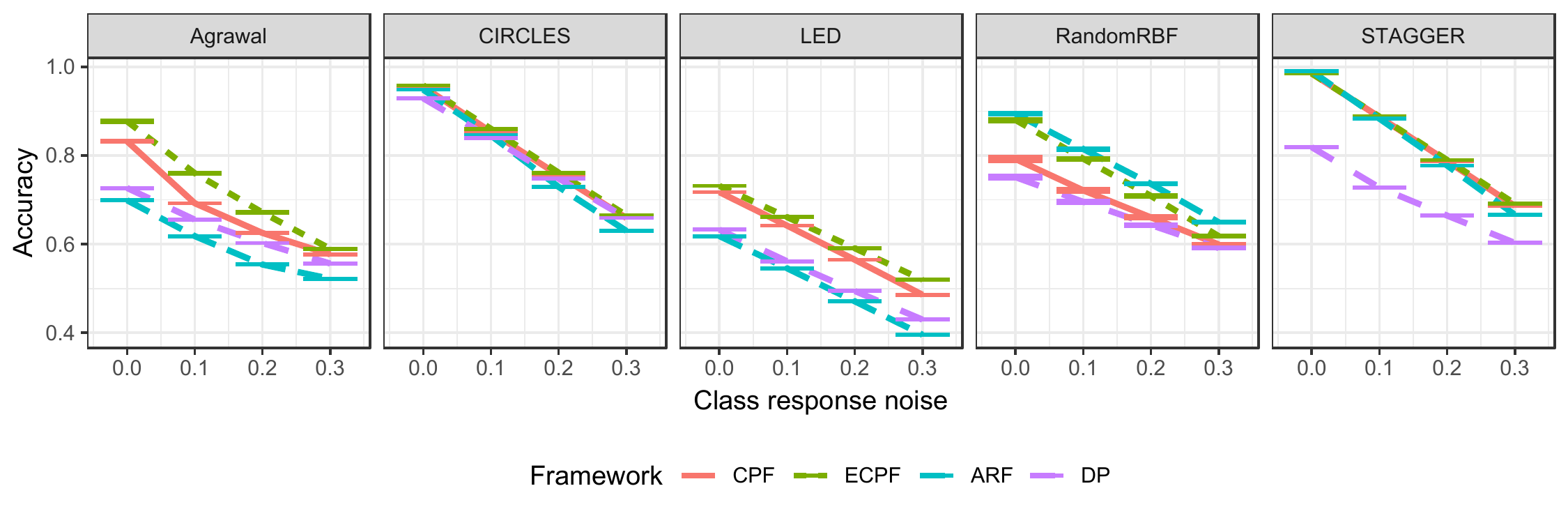}
	\caption{Accuracy on synthetic datasets with noise applied to response class}
	\label{plotSynthRespNoise}
\end{figure}

Fig. \ref{plotSynthNoise} show results for experiments run on datasets with added noise. We applied noise from a Gaussian distribution $1 + \mathcal{N}(\mu = 0, \theta = attnoise/3),$ $attnoise \in \{0.0, 0.1, 0.25, 0.5\}$ multiplicatively to all numeric predictor (attribute) variables in our datasets. For categorical variables, we applied an $attnoise$ chance that the variable would be switched with another random but valid value. Comparing circumstances with no noise to the most noise (\emph{i.e} $attnoise = 0.5$), ARF's accuracy was most affected on CIRCLES, RandomRBF and STAGGER. CPF was most affected on Agrawal, and ECPF was most affected on LED. However, on LED with $attnoise = 0.5$, no technique outperformed guessing the class randomly. Across experiments, attribute noise seemed to effect all techniques fairly evenly, with DP appearing the most robust in its presence. Despite this, DP still was less accurate than ECPF on all datasets at all level of noise, except for LED with $attnoise = 0.5$.

Fig. \ref{plotSynthRespNoise} show results for experiments run on datasets with noise applied to the response or class variable. We randomly changed the class variable to another based on random chance (of $0, 0.1, 0.2$ or $0.3$). As our techniques reuse classifiers based on similar classifications, noise in the class variable could affect when classifiers are succesfully reused. As would be expected, all techniques dropped in accuracy when the class variable was randomly changed.  Generally the best performing techniques were affected the most by response noise. ECPF dropped the most on Agrawal and RandomRBF and CPF dropped the most on LED. ARF's accuracy was affected the most on CIRCLES and STAGGER. DP was affected the least across datasets, but is still consistently the least accurate.

\begin{figure}[!htbp]
	\centering
	\includegraphics[width=0.9\textwidth]{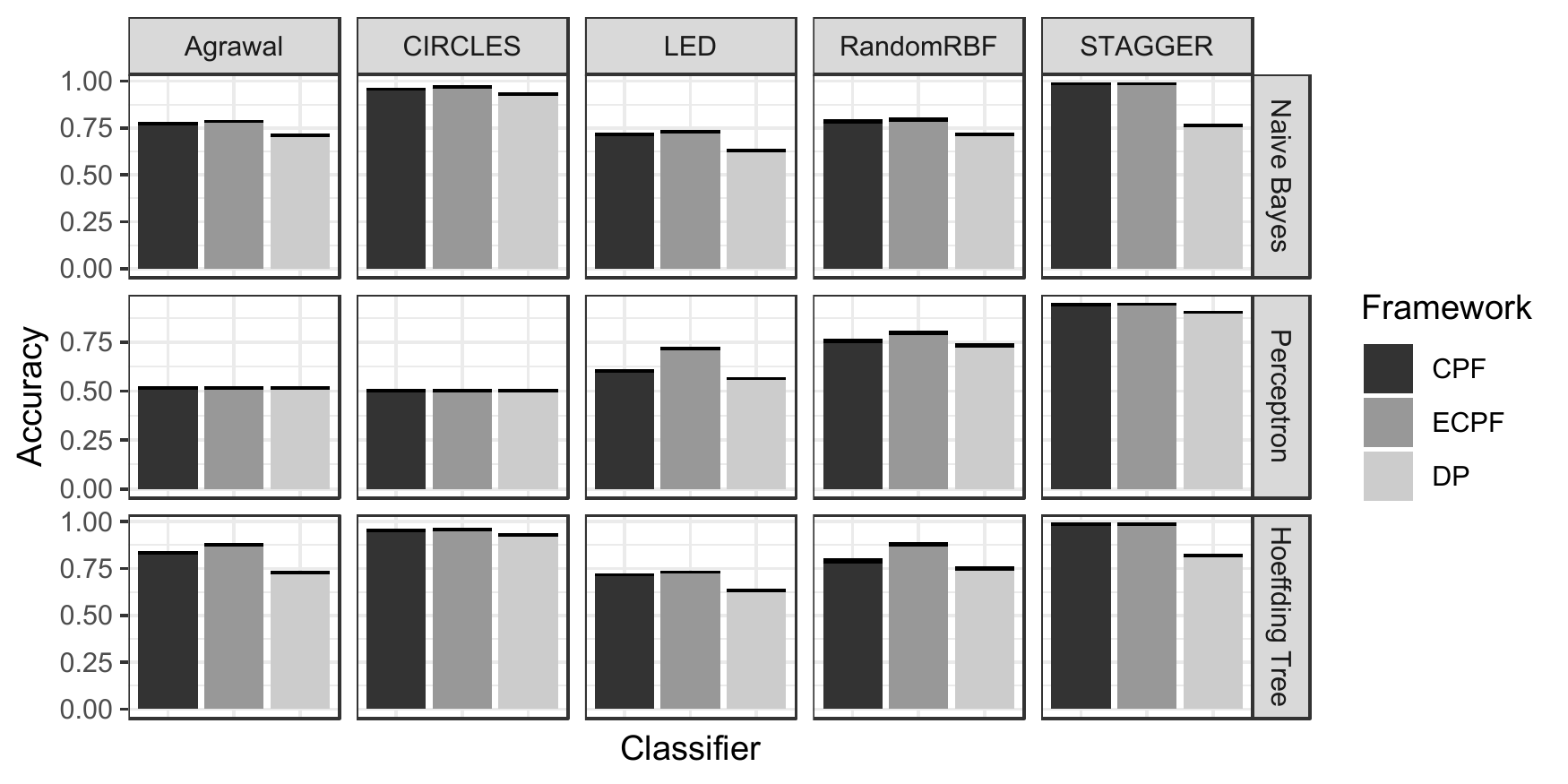}
	\caption{Accuracy on synthetic datasets across classifier types by framework}
	\label{plotSynthClassAcc}

	\includegraphics[width=0.9\textwidth]{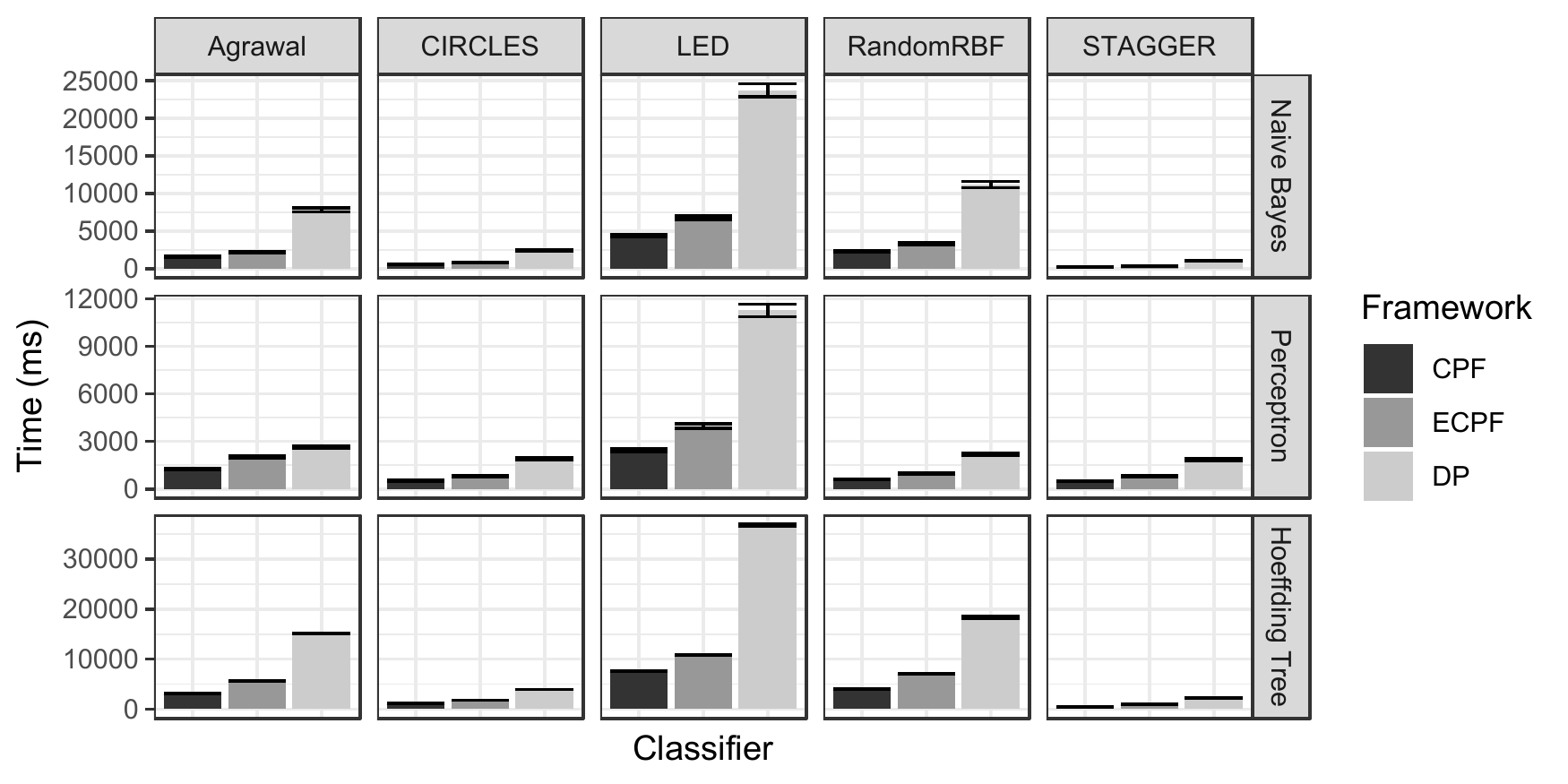}
	\caption{Runtime on synthetic datasets across classifier types by framework}
	\label{plotSynthClassTime}
\end{figure}

In Fig. \ref{plotSynthClassAcc}, we show accuracy for ECPF compared to CPF and DP with different classifier types used within the frameworks (ARF is reliant on its Hoeffding Tree-based classifier so cannot be used with others). The implementations of these classifier types are also from MOA, with Hoeffding Tree used for all other experiments. CPF and ECPF were never less accurate than DP. However, for two datasets with a Perceptron classifier (Agrawal and CIRCLES), the same accuracy was achieved by all frameworks. This may be due to the classifier's limited capacity to capture class boundaries. In other experiments, CPF and ECPF generally maintained their improvement above DP. As per Fig. \ref{plotSynthClassTime}, relative speed of frameworks are generally not dependent on the classifier used. 

Across these experiments, we show that CPF and ECPF maintain their performance and are generally no more affected by permutations of underlying data nor classifier type used than other state-of-the-art techniques. These results support them both as robustly performing frameworks on datasets with recurring drifts.

\subsection{Experiments exploring the behaviour of ECPF}

In this section, we explore how our proposed framework is affected by drift detector choice and parameters. Specifically, we explore how varying drift detector affects classification accuracy, runtime and drift detection. We explore how changing the fade points parameter $f$ and the similarity parameter $m$ affect each framework. These experiments provide insight into how our techniques behave dependent on their configuration.

	
	
	

\begin{table}
	\scriptsize
	\caption{Accuracy, runtime and drifts detected across synthetic datasets with different drift detectors (400 drifts per dataset)}
	\label{tabDetectors}
	\begin{tabular}{p{1.35cm}cccccccc}
		\toprule
		Stream & \shortstack{Drift \\ detector} & Framework & Acc & $\sigma$ & \shortstack{Time \\ (s)} & $\sigma$ & \shortstack{Drifts \\ detected} & $\sigma$\\
		\midrule
			Agrawal & PerfectDetector & ECPF & 87.7\% & 0.8\% & 5.6 & 0.2 & 400 & 0\\
			Agrawal & PerfectDetector & CPF & 83.2\% & 0.5\% & 3.0 & 0.1 & 400 & 0\\
			Agrawal & HDDM-A & ECPF & 82.2\% & 1.1\% & 5.7 & 0.2 & 384 & 16\\
			Agrawal & HDDM-A & CPF & 78.2\% & 0.4\% & 3.6 & 0.1 & 433 & 13\\
			Agrawal & MagSeed & ECPF & 79.2\% & 1.3\% & 4.4 & 0.3 & 930 & 97\\
			Agrawal & MagSeed & CPF & 70.7\% & 0.4\% & 3.7 & 0.1 & 1013 & 40\\
			Agrawal & RDDM & ECPF & 83.9\% & 1.0\% & 6.1 & 0.4 & 588 & 55\\
			Agrawal & RDDM & CPF & 76.7\% & 0.3\% & 4.1 & 0.1 & 738 & 43\\
			\hline
			\addlinespace
			CIRCLES & PerfectDetector & ECPF & 95.8\% & 0.1\% & 1.8 & 0.0 & 400 & 0\\
			CIRCLES & PerfectDetector & CPF & 95.6\% & 0.4\% & 1.1 & 0.0 & 400 & 0\\
			CIRCLES & HDDM-A & ECPF & 93.4\% & 0.1\% & 2.1 & 0.0 & 167 & 11\\
			CIRCLES & HDDM-A & CPF & 93.3\% & 0.3\% & 1.4 & 0.0 & 182 & 17\\
			CIRCLES & MagSeed & ECPF & 94.2\% & 0.5\% & 1.5 & 0.0 & 689 & 56\\
			CIRCLES & MagSeed & CPF & 93.1\% & 0.2\% & 1.0 & 0.0 & 618 & 24\\
			CIRCLES & RDDM & ECPF & 95.5\% & 0.2\% & 2.0 & 0.1 & 809 & 161\\
			CIRCLES & RDDM & CPF & 94.9\% & 0.2\% & 1.2 & 0.0 & 512 & 36\\
			\hline
			\addlinespace
			LED & PerfectDetector & ECPF & 73.1\% & 0.1\% & 10.7 & 0.2 & 400 & 0\\
			LED & PerfectDetector & CPF & 71.7\% & 0.0\% & 7.5 & 0.1 & 400 & 0\\
			LED & HDDM-A & ECPF & 72.1\% & 0.1\% & 10.6 & 0.2 & 363 & 10\\
			LED & HDDM-A & CPF & 70.8\% & 0.1\% & 7.5 & 0.1 & 372 & 5\\
			LED & MagSeed & ECPF & 71.4\% & 0.3\% & 10.6 & 0.3 & 993 & 87\\
			LED & MagSeed & CPF & 61.1\% & 0.2\% & 11.5 & 0.3 & 1607 & 43\\
			LED & RDDM & ECPF & 72.7\% & 0.1\% & 11.2 & 0.2 & 462 & 11\\
			LED & RDDM & CPF & 71.1\% & 0.1\% & 7.8 & 0.1 & 400 & 4\\
			\hline
			\addlinespace
			RandomRBF & PerfectDetector & ECPF & 88.0\% & 1.4\% & 7.0 & 0.1 & 400 & 0\\
			RandomRBF & PerfectDetector & CPF & 79.2\% & 2.7\% & 4.1 & 0.1 & 400 & 0\\
			RandomRBF & HDDM-A & ECPF & 83.1\% & 1.7\% & 6.7 & 0.1 & 374 & 27\\
			RandomRBF & HDDM-A & CPF & 78.1\% & 2.7\% & 4.3 & 0.1 & 381 & 32\\
			RandomRBF & MagSeed & ECPF & 77.2\% & 2.6\% & 5.8 & 0.2 & 838 & 79\\
			RandomRBF & MagSeed & CPF & 75.2\% & 2.7\% & 4.5 & 0.1 & 757 & 49\\
			RandomRBF & RDDM & ECPF & 85.4\% & 1.5\% & 7.1 & 0.1 & 484 & 60\\
			RandomRBF & RDDM & CPF & 78.0\% & 2.7\% & 4.4 & 0.1 & 460 & 46\\
			\hline
			\addlinespace
			STAGGER & PerfectDetector & ECPF & 98.5\% & 0.0\% & 1.1 & 0.1 & 400 & 0\\
			STAGGER & PerfectDetector & CPF & 98.5\% & 0.0\% & 0.4 & 0.0 & 400 & 0\\
			STAGGER & HDDM-A & ECPF & 99.7\% & 0.0\% & 1.3 & 0.1 & 401 & 1\\
			STAGGER & HDDM-A & CPF & 98.3\% & 0.0\% & 0.7 & 0.0 & 400 & 0\\
			STAGGER & MagSeed & ECPF & 98.8\% & 0.2\% & 1.4 & 0.1 & 1052 & 89\\
			STAGGER & MagSeed & CPF & 98.4\% & 0.1\% & 0.6 & 0.0 & 829 & 21\\
			STAGGER & RDDM & ECPF & 99.8\% & 0.0\% & 1.2 & 0.1 & 495 & 64\\
			STAGGER & RDDM & CPF & 98.5\% & 0.0\% & 0.5 & 0.0 & 421 & 13\\
		\bottomrule
	\end{tabular}
\end{table}

Table \ref{tabDetectors} show results of experiments using ECPF with different drift detectors on datasets. PerfectDetector refers to our default approach to other experiments - detecting a drift 60 instances after it occurs on each dataset. Each stream had 400 real drifts occurring. In terms of accuracy, we see that our CPF and ECPF are affected similarly by drift detector choice. Generally, the PerfectDetector yields the best accuracy and MagSeed the worst, with similar accuracy between RDDM and HDDM-A. Both variants of CPF generally were more accurate when drift detectors did not detect too many drifts. The clearest example of too many drifts being detrimental for accuracy is for CPF with MagSeed on LED. We can see by the drifts detected that this is likely due to a high false positive rate of drift detection for this combination, with a greater effect than when using ECPF. As LED has ten classes, conceptual equivalence may lead to fewer correctly reused classifiers for CPF, resulting in a drop in accuracy and more drifts detected. However, the same pattern is not seen with HDDM-A nor RDDM, suggesting that it may be related to the different approaches to drift detection that MagSeed takes: it compares windows of errors rather than error rate. This suggests the ECPF is better equipped to maintain accuracy in the presence of false positive drifts than CPF, likely due to having a 'new' classifier it can revert to when the reused classifier is underperforming - drift will only be detected when the best performing of the new and reused classifier drop in accuracy.

RDDM can have high false positive rates of drift, but this does not seem to negatively impact accuracy when compared to HDDM-A. This aligns with the findings about RDDM in \cite{bar2018}. HDDM-A seems to work effectively in conjunction with our techniques, rarely being too sensitive in its drift detection. One exception is for CIRCLES, where it is possible that the change between some concepts does not result in sufficient enough classifier error to trigger drift detection. Runtime rarely is dramatically affected by drift detector choice, though in the case of LED and CPF, it appears that many false positive drifts detected can lead to CPF running slower than ECPF.

\begin{figure}[!htbp]
	\includegraphics[width=\textwidth]{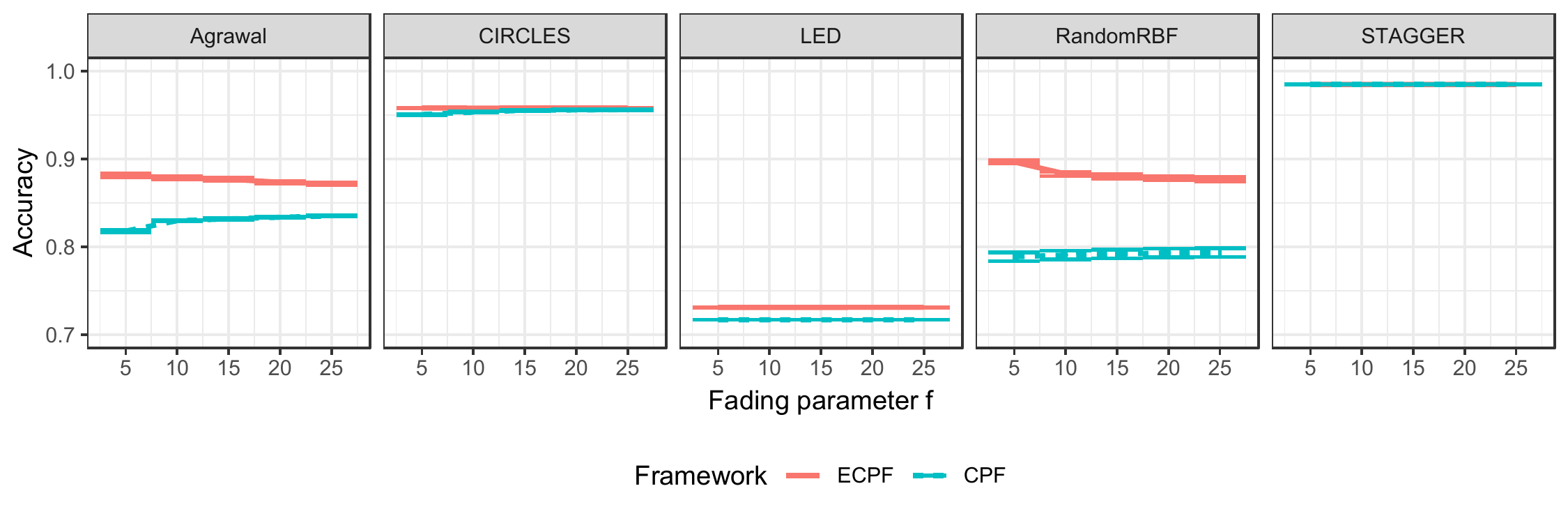}
	\caption{Accuracy on synthetic datasets with varied fading parameter $f$}
	\label{plotSynthFading}
\end{figure}

In Fig. \ref{plotSynthFading}, we show results for ECPF with varied fading points $f$. The higher this value is, the greater number of classifiers our framework is likely to hold at any time, and the longer they can take before selecting which classifiers to maintain. Interestingly, on two of five datasets, ECPF benefits from reduced $f$, likely keeping fewer classifiers. Meanwhile on three of five datasets CPF benefits from likely keeping more classifiers. ECPF copies a classifier before reusing it, so when it incorrectly reuses a classifier, the original is intact. This is not the case for CPF. CPF likely benefits from a greater variety of classifiers as it has more chance of reusing a better classifier when it has a wide range to select from. This also avoids inappropriately reusing the wrong classifier, which could lead to training it on instances from a distinct concept. ECPF can likely benefit by reusing and training classifiers repeatedly when concept recur; these datasets are limited to no more than five distinct concepts. This means that each classifier becomes a better expert at its own concept, and there is no danger of losing that classifier through inappropriate reuse. Based on these results, CPF is likely to give better performance with higher $f$. ECPF appears to benefit from lower $f$ but in actuality, this is likely to depend on the number of true distinct concepts in the underlying dataset. This is why we selected $f = 15$ for real-world experiments and would recommend this as the default choice unless it is known that there is a limited number of underlying concepts in a data stream.

\begin{figure}[!htbp]
	\includegraphics[width=\textwidth]{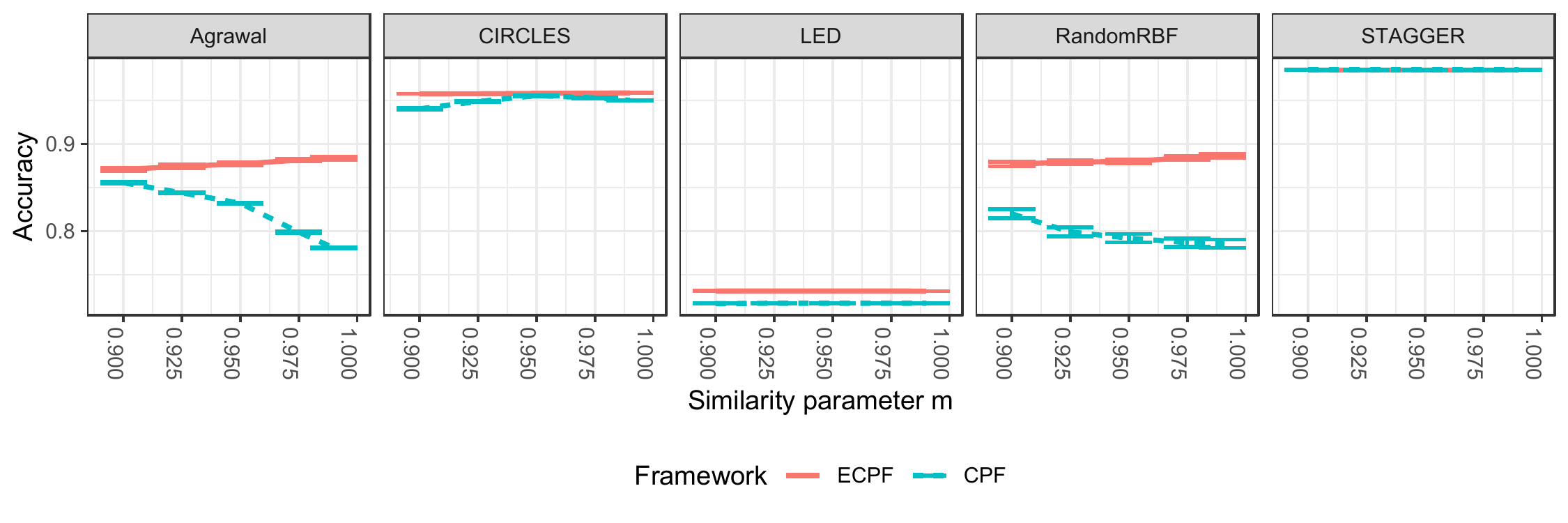}
	\caption{Accuracy on synthetic datasets with varied similarity parameter $m$}
	\label{plotSynthSim}
\end{figure}

In Fig. \ref{plotSynthSim}, we show results for synthetic datasets when varying similarity parameter $m \in \{0.900, 0.925. 0.950, 0.975, 0.990\}$. This parameter clearly impacts ECPF on two datasets: higher $m$ leads to better performance on Agrawal and RandomRBF for ECPF while the converse is true for CPF. Meanwhile, $m=0.95$ appears to be the best setting for CPF on CIRCLES. The similarity parameter determines either how accurate or how similar to an untrained classifier an existing classifier must be for reuse. It also determines how similar two classifiers must be before removing one classifier and keeping the other to represent it. As ECPF creates a new classifier on every drift, it will more regularly consider $m$ when finding classifiers to represent one another than CPF generally will. High similarity sets a higher threshold for CPF to reuse classifiers while ECPF will always reuse a classifier after every drift. CPF likely loses accuracy on harder problems with higher $m$ because of reduced classifier reuse. ECPF is likely penalised by lower $m$ as the threshold required to see two classifiers as similar is lower. This could result, for example, in two classifiers trained on distinct concepts each being removed in favour of one classifier that approximates both concepts. Higher $m$ makes this type of error less likely.

\section{Conclusion and future work}
\label{conclusion}

Through this paper we have detailed the Enhanced Concept Profiling Framework that can be used to accurately and quickly classify data streams with recurring drifts. Our approach provides accurate classification of data streams through recognising recurring drifts by identifying conceptually equivalent classifiers. By recognising where classifiers are similar, our technique can reuse classifiers trained on an equivalent concept. ECPF maintains a new classifier and a reused classifier after each drift, and on the next dift detection, it can select which is the more accurate classifier to store for future use. Through copying classifiers each time it reuses them, ECPF can train a classifier on a concept which may or may not be recurring without losing the original classifier.

We have shown that ECPF is significantly more accurate than state-of-the-art frameworks such as Adaptive Random Forests (ARF) on synthetic data streams with recurring concepts. ECPF is robust, performing at least as consistently as state-of-the-art approaches in the presence of frequent drift, varied class balance, different types of noise and different types of classifier. On real-world datasets we show that ECPF is at least as accurate as a state-of-the-art classifier reuse fraemwork. In addition, ECPF is not significantly less accurate than state-of-the-art ensemble frameworks. In the worst case, ECPF was only $4\%$ less accurate than Adaptive Random Forests across real-world datasets tested. Most importantly, we have shown ECPF is significantly faster than ensemble techniques and generally much faster than Diversity Pool (DP). ECPF took under a fifth of the combined time to classify real-world datasets as DP, and under a sixth of the time it took ARF.

We have shown that classifier reuse frameworks are competitive with ensemble approaches while capable of running much more quickly than the latest ensemble techniques. However, ECPF only outperforms a state-of-the-art ensemble approach on synthetic datasets with abrupt concept drift and distinct concepts. Finding a way to handle gradual concept drift with a classifier reuse framework would be an achievement that could allow fast stream classification at a level of performance rivalling the best ensembles. Extensions of this framework to work with regression problems with recurrent drift would be interesting and make it applicable to many other areas. Finally, approaches that could proactively predict drift occurence and upcoming concepts based on past drifts and concepts seen could lead to more dramatic improvements in accuracy for our technique.

\clearpage
\bibliography{ESA_ECPF}

\end{document}